\documentclass{article}

\usepackage{microtype}
\usepackage{graphicx}
\usepackage{subcaption}
\usepackage{booktabs} 
\usepackage{textcomp} 
\usepackage{gensymb}
\usepackage{wrapfig}
\usepackage{enumitem}
\usepackage{dblfloatfix}    
\usepackage[toc,page]{appendix}
\usepackage[all]{nowidow}  

\usepackage[hidelinks]{hyperref}

\usepackage[accepted]{icml2020}

\renewcommand{\paragraph}[1]{\textbf{#1}\quad}

\icmltitlerunning{Automatic Shortcut Removal for Self-Supervised Representation Learning}

\usepackage{amsmath}

\DeclareMathOperator*{\argmin}{arg\,min}

\newcommand{\relpatchloc}{\emph{Relative patch location}}
\newcommand{\jigsaw}{\emph{Jigsaw}}
\newcommand{\rotation}{\emph{Rotation}}
\newcommand{\exemplar}{\emph{Exemplar}}

\newcommand{\cifar}{\emph{CIFAR-10}}
\newcommand{\imagenet}{\emph{ImageNet}}
\newcommand{\places}{\emph{Places205}}
\newcommand{\youtube}{\emph{YouTube1M}}

\newcommand{\reclossscale}{\lambda}
\newcommand{\lens}{L}
\newcommand{\featureextractor}{F}

\newcommand{\supplement}{appendix}

\begin{document}

\twocolumn[
\icmltitle{Automatic Shortcut Removal for Self-Supervised Representation Learning}



\icmlsetsymbol{equal}{*}

\begin{icmlauthorlist}
\icmlauthor{Matthias Minderer}{g,r}
\icmlauthor{Olivier Bachem}{g}
\icmlauthor{Neil Houlsby}{g}
\icmlauthor{Michael Tschannen}{g}
\end{icmlauthorlist}

\icmlaffiliation{g}{Google Research, Brain Team, Z\"{u}rich, Switzerland}

\icmlaffiliation{r}{Work done as part of the Google AI Residency}

\icmlcorrespondingauthor{Matthias Minderer}{mjlm@google.com}

\icmlkeywords{Machine Learning, ICML, representation learning, self-supervised learning}

\vskip 0.3in
]



\printAffiliationsAndNotice{}  


\begin{abstract}
In self-supervised visual representation learning, a feature extractor is trained on a ``pretext task" for which labels can be generated cheaply, without human annotation. A central challenge in this approach is that the feature extractor quickly learns to exploit low-level visual features such as color aberrations or watermarks and then fails to learn useful semantic representations. Much work has gone into identifying such ``shortcut" features and hand-designing schemes to reduce their effect. Here, we propose a general framework for mitigating the effect shortcut features. Our key assumption is that those features which are the first to be exploited for solving the pretext task may also be the most vulnerable to an adversary trained to make the task harder. We show that this assumption holds across common pretext tasks and datasets by training a ``lens" network to make small image changes that maximally reduce performance in the pretext task. Representations learned with the modified images outperform those learned without in all tested cases. Additionally, the modifications made by the lens reveal how the choice of pretext task and dataset affects the features learned by self-supervision.
\end{abstract}
\section{Introduction}

\begin{figure}[ht]
\begin{center}
    \vspace{5mm}
    \centerline{\includegraphics[width=\columnwidth,trim={0in 1in 0in 0in},clip]{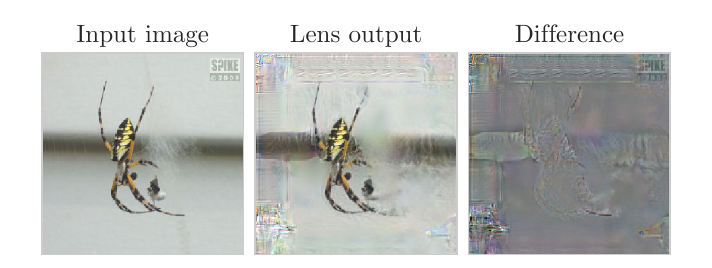}}
    \vskip -0.1in
    \caption{
    Example of automatic shortcut removal for the \rotation{} prediction pretext task. The lens learns to remove features that make it easy to solve the pretext task (concretely, it conceals watermarks in this example). Shortcut removal forces the network to learn higher-level features to solve the pretext task and improves representation quality.
    }
    \label{fig:figure_1_example_images}
\end{center}
\end{figure}
In self-supervised visual representation learning, a neural network is trained using labels that can be generated cheaply from the data rather than requiring human labeling. These artificial labels are used to create a ``pretext task'' that ideally requires learning abstract, semantic features useful for a large variety of vision tasks. A network pre-trained on the pretext task can then be transferred to other vision tasks for which labels are more expensive to obtain, e.g. by learning a head or fine-tuning the network for the target task. 

Self-supervision has recently led to significant advances in unsupervised visual representation learning, with the first self-supervised methods outperforming supervised ImageNet pre-training on selected vision benchmarks \cite{henaff2019data,he2019momentum,misra2019self}. 

Yet, defining sensible pretext tasks remains a major challenge because neural networks are biased towards exploiting the simplest features that allow solving the pretext task. This bias works against the goal of learning semantically meaningful representations that transfer well to a wide range of target tasks. Simple solutions to pretext tasks can be unintuitive and surprising. For example, in the self-supervised task of predicting the orientation of rotated training images \cite{gidaris2018unsupervised}, logos and watermarks allow the network to predict the orientation of the input image by learning simple text features, rather than transferable object representations (Figure~\ref{fig:figure_1_example_images}). Similarly, nearly imperceptible color fringes introduced by chromatic aberrations of camera lenses provide a signal for context-based self-supervised methods that is strong enough to significantly reduce the quality of representations learned from these tasks, unless they are specifically addressed by augmentation schemes \cite{doersch2015unsupervised, noroozi2016unsupervised}.

Many such data augmentation procedures have been proposed, but have relied on the intuition and creativity of researchers to identify shortcut features. We aim to break this pattern and propose a simple method to remove shortcuts automatically. The key insight underlying our method is that visual features which allow a network to easily solve a pretext task may also be features which an adversary can easily exploit to make the task harder. We therefore propose to process images with a lightweight image-to-image translation network, called ``lens'' (borrowing the terminology from \citealp{sajjadi2018tempered}), which is trained adversarially to reduce performance on the pretext task without deviating much from the original image.\footnote{We address the unwanted removal of potentially useful features in Section~\ref{sec:shortcut-def}.} Once trained, the lens can be applied to unseen images, so it can be used downstream when transferring the network to a new task. We show that the lens leads to significant improvements across a variety of pretext tasks and datasets. Furthermore, the lens can be used to visualize the shortcuts by inspecting the difference image between the input and the output images. The changes made to the image by the lens provide insights into how shortcuts differ across tasks and datasets, which we use to make suggestions for future task design.

In summary, we make the following contributions:

\begin{itemize}[leftmargin=12pt, topsep=0pt, partopsep=0pt, itemsep=0pt, parsep=0pt]
    \item We propose a simple and general method for automated removal of shortcuts which can be used with virtually any pretext task.
    \item We validate the proposed method on a wide range of pretext tasks and on two different training datasets (ImageNet and YouTube-8M frames), showing consistent improvements across all methods, upstream training datasets, and two downstream/evaluation datasets (ImageNet and Places205). In particular, our method can replace preprocessing procedures that were hand-engineered to remove shortcuts.
    \item We use the lens to compare shortcuts across different pretext tasks and data sets. This analysis provides useful insights into data set and pretext-specific shortcuts.
\end{itemize}

\section{Related work}

\paragraph{Self-supervised learning.} Self-supervised learning (SSL) has attracted more and more interest in the computer vision and machine learning community over the past few years. Early approaches (pretext tasks) involve exemplar classification \cite{dosovitskiy2014discriminative}, predicting the relative location of image patches \cite{doersch2015unsupervised}, solving jigsaw puzzles of image patches \cite{noroozi2016unsupervised}, image colorization \cite{zhang2016colorful}, object counting \cite{noroozi2017representation}, predicting the orientation of images \cite{gidaris2018unsupervised}, and clustering \cite{caron2018deep}. These methods are typically trained on ImageNet and their performance is evaluated by training a linear classification head on the frozen representation.

The challenge of avoiding trivial "shortcut" solutions to pretext tasks was first discussed by \citet{doersch2015unsupervised}, who described how chromatic aberrations and matching of patterns across boundaries act as shortcuts for predicting the relative location of image patches. Since then, new pretext tasks are usually proposed along with procedures to mitigate the effect of shortcuts. Common shortcuts include color aberrations \cite{doersch2015unsupervised}, re-sampling artifacts \cite{noroozi2017representation}, compression artifacts \cite{wei2018learning}, and salient lines or grid patterns \cite{doersch2015unsupervised, noroozi2016unsupervised}. To counter the effects of these shortcuts, various pre-processing strategies have been developed, for example channel dropping \cite{doersch2015unsupervised, doersch2017multi}, channel replication \cite{lee2017unsupervised}, conversion to gray scale \cite{noroozi2017representation}, chroma blurring \cite{mundhenk2018improvements}, and spatial jittering \cite{doersch2015unsupervised, mundhenk2018improvements}. Recently, \citet{jenni2018artifacts} proposed a pretext task based on detecting synthetic artifacts. To remove shortcut solutions to this task and improve the learned representations, they adversarially trained a ``repair network", which is conceptually related to our lens network. Our work generalizes this approach to arbitrary pretext tasks.

One of the premises of SSL is that it can be applied to huge data sets for which human labeling would be too expensive. \citet{goyal2019scaling} explore this aspect and find that large-scale SSL can outperform supervised pretraining.
Another research direction is to combine several pretext tasks into one, which often improves performance \cite{doersch2017multi, feng2019self, misra2019self}. Using multiple pretext tasks could reduce the effect of shortcuts that are not shared across tasks. These efforts may benefit from our comparison of shortcuts across tasks.

More recently, contrastive methods gained popularity \cite{oord2018representation, hjelm2018learning, bachman2019learning, tian2019contrastive, henaff2019data, he2019momentum, chen2020simple}. These methods are based on the principle of learning a representation which associates different views (e.g. different augmentations) of the same training example with similar embeddings, and views of different training examples with dissimilar embeddings \cite{tschannen2020mutual}. 
There are numerous parallels between pretext task-based and contrastive methods \cite{he2019momentum}, and our method in principle applies to both types of approaches.

\paragraph{Adversarial training.}
Our method is related to adversarial training and we give a brief overview of works relevant for this paper (see \citet{akhtar2018threat} for a survey). Adversarial examples are small, imperceptible perturbations to the input of a classifier that lead to a highly confident misclassification of the input \cite{szegedy2013intriguing}. Deep neural networks are particularly susceptible to adversarial perturbations. A plethora of adversarial training techniques were proposed to make them more robust, e.g. the fast gradient sign method (FGSM; \citealp{goodfellow2014explaining}) or the projected gradient descent defense \cite{madry2017towards}. 

Adversarial training can significantly improve semi-supervised learning when combined with label propagation \cite{miyato2018virtual}. However, only very recently \citet{xie2019adversarial} succeeded in developing an adversarial training procedure that substantially improves classification accuracy on unperturbed images in the context of (fully) supervised learning. Somewhat related, \citet{santurkar2019image} present evidence that adversarially trained classifiers learn more abstract semantic features. 
We emphasize that all these works use hand-annotated ground-truth labels carrying much more information than pretext labels, and we believe that adversarial training has an even higher potential for SSL.
\section{Method}

\begin{figure}[t]
\begin{center}
    \centerline{\includegraphics[width=\columnwidth]{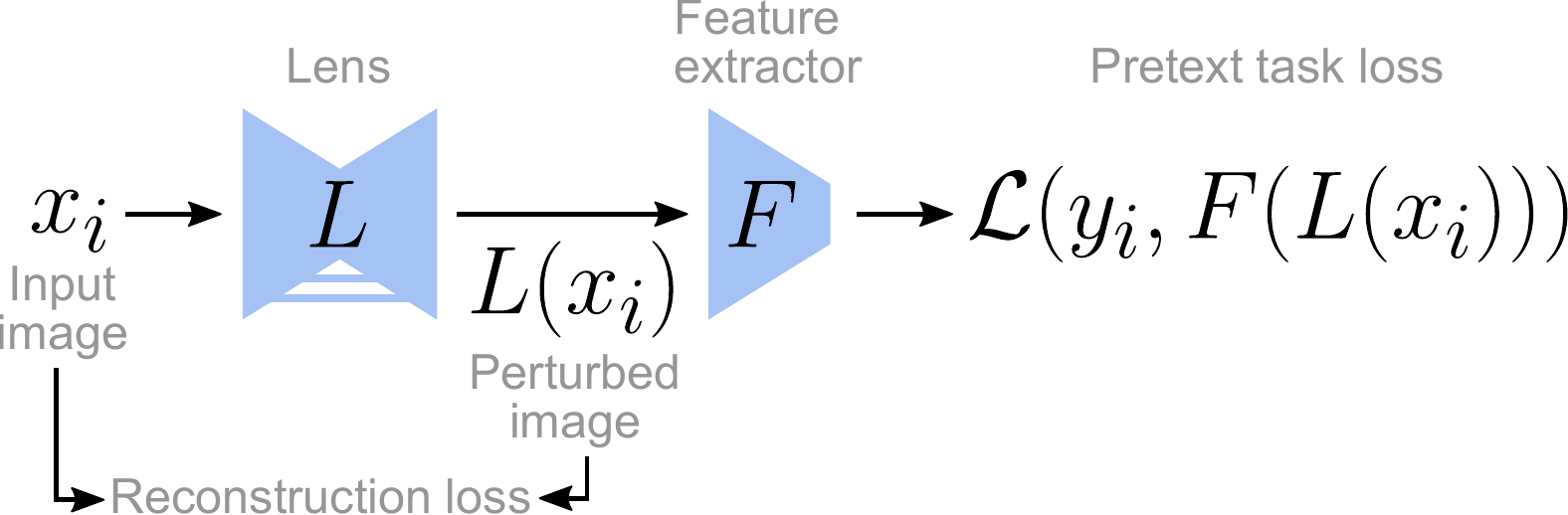}}
    \caption{Model schematic. For the experiments in this paper, we use the \emph{U-Net} architecture for the lens $L$ and the \emph{ResNet50 v2} architecture for the feature extractor $F$. We use an $\ell_2$ loss as the reconstruction loss for simplicity, but other choices are possible.}
    \label{fig:model_schematic}
    \vspace{-5mm}
\end{center}
\end{figure}

We propose to improve self-supervised visual representations by processing images with a lightweight image-to-image translation network, or ``lens'', that is trained adversarially to reduce performance of the feature extractor network on the pretext tasks. In this section, we outline our approach in more detail. We start by defining the notion of ``shortcut'' visual features for the purpose of this study.

\subsection{What are shortcuts?} \label{sec:shortcut-def}

Shortcuts have been described as ``trivial solutions'' to the pretext task that must be avoided to ``ensure that the task forces the network to extract the desired information'' \cite{doersch2015unsupervised}. In other words, shortcuts are easily learnable features that are predictive of the pretext label, and allow the network to stop learning once found. As described below, shortcuts can be identified by using an adversarial loss to learn which visual features allow solving the pretext task easily.

It is tempting to think of shortcuts exclusively as useless artefactual visual features, because the most prominent examples fall in this category (e.g. watermarks and chromatic aberrations). However, the downstream tasks are typically assumed to be unknown in the transfer learning scenario. It is therefore impossible to know \emph{a priori} whether a shortcut feature is undesired and can be safely removed (such as the watermark in Figure~\ref{fig:figure_1_example_images}, if the downstream task is \imagenet{} classification), or is useful downstream despite being an easy solution to the pretext task (such as the eyes of the dogs in Figure~\ref{fig:reconstruction_loss_scale_sweep}). In fact, for any potential shortcut, a downstream task could be conceived that depends on this feature (e.g. watermark detection). 

We therefore think about shortcuts purely in terms of the pretext task, rather than in terms of their usefulness downstream. Since we cannot know in general whether it is safe to completely remove a shortcut feature, a general approach to shortcut mitigation should encourage the network to learn non-shortcut features \emph{and} shortcuts. We achieve this by providing both lensed and non-lensed images during pretraining, and then combining representations obtained with and without shortcut removal before use in downstream tasks. This ensures that the automatic shortcut removal never reduces the information present in the representations. We provide empirical justification for this approach in Section~\ref{subsec:results}.

\subsection{Automatic adversarial shortcut removal}

We start by formalizing the common setup for pretext task-based SSL, and then describe how we modify this setup to prevent shortcuts. 

In pretext task-based SSL, a neural network $F$ (sometimes also called ``encoder'' or ``feature extractor'') is trained to predict machine-generated pretext targets $y_i$ from inputs $x_i$. The pretext task is learned by minimizing a loss function $\mathcal{L}_{\text{SSL}}$ which usually takes the form $\mathcal{L}_{\text{SSL}} = \sum_{i=1}^N L_\text{SSL} (F(x_i), y_i)$, where $N$ is the number of training examples and $L_\text{SSL}$ is often a cross-entropy loss.

To remove shortcuts, we introduce a lens network $L$ that (slightly) modifies its inputs $x_i$ and maps them back to the input space, before feeding them to the representation network $F$ (Figure~\ref{fig:model_schematic}). When using the lens, the pretext task loss becomes $\mathcal{L}_{\text{SSL}} = \sum_{i=1}^N L_\text{SSL} (F(L(x_i)), y_i)$ and $F$ is trained to minimize this loss, as before. As motivated in Section~\ref{sec:shortcut-def}, we train the lens adversarially  against $\mathcal{L}_{\text{SSL}}$ to increase the difficulty of the pretext-task. We consider two loss variants that were previously considered in the adversarial training literature \cite{kurakin2016adversarial}: \emph{full} and \emph{least likely}. 

The \emph{full} adversarial loss is simply the negative task loss: $\mathcal{L}_{\text{adv}} = -\mathcal{L}_{\text{SSL}}$.
This type of adversarial training is applicable to any pretext task loss.

For classification pretext tasks, we can alternatively train the lens to bias the predicted class probabilities towards the \emph{least likely} class. The loss hence becomes:
\begin{align*}
\mathcal{L}_{\text{adv}} &= \sum_{i=1}^N L_\text{SSL} (F(L(x_i)), y_i^\text{LL}), \quad \text{where} \\
y_i^\text{LL} &= \argmin_y p(y|F(L(x_i))).
\end{align*}
The lens is also trained with a reconstruction loss to avoid trivial solutions: $\mathcal{L}_{\text{lens}} = \mathcal{L}_{\text{adv}} + \lambda \mathcal{L}_{\text{rec}}$, where $\mathcal{L}_{\text{rec}} = \sum_{i=1}^N \Vert x_i - L(x_i)\Vert_2^2$ is a pixel-wise $\ell_2$ reconstruction loss and $\reclossscale > 0$ is a hyperparameter that trades off the strength of the adversarial attack against reconstruction quality.  

\subsection{Hyperparameters and design choices}

Before presenting a comprehensive experimental evaluation of the proposed method, we first discuss major hyperparameters and design choices, and compare our method to standard adversarial training.

\paragraph{Reconstruction loss.} We use an $\ell_2$ loss for $\mathcal L_\text{rec}$ due to its simplicity and stability. Other choices are possible and interesting, in particular losses going beyond pixel-wise similarity, measuring semantic similarity such as feature matching \cite{salimans2016improved} or perceptual losses \cite{zhang2018unreasonable}. However, note that these losses themselves often require supervised pretraining, hence defeating the purpose of (unsupervised) SSL. We discuss the effect of $\mathcal L_\text{rec}$ in the context of different pretext tasks in Section~\ref{subsec:comparison_across_tasks}.

\paragraph{Selection of $\lambda$.} The reconstruction loss scale $\reclossscale$ is the most important hyperparameter for lens performance and the only one that we vary between tasks. The optimal value for $\reclossscale$ depends primarily on the scale of the pretext task loss, but also on the dataset and data augmentation applied prior to feeding the examples to the lens.

\paragraph{Network architectures.} For the feature extraction network $\featureextractor$, we use the default \emph{ResNet50 v2} architecture \cite{he2016identity} (i.e. with a channel widening factor of 4 and a representation size of 2048 at the \emph{pre-logits} layer) unless otherwise noted. For the lens $\lens$, we use a variant of the \emph{U-Net} architecture \cite{ronneberger2015unet}. The encoder and decoder are each a stack of $4$ residual blocks (ResNet50 v2 residual blocks) with $2\times$ down-sampling and up-sampling, respectively, after each block (a complete description can be found in the supplementary material). This lens architecture has 3.87M parameters, less than one sixth of the \emph{ResNet50 v2} network used to compute the representation (23.51M).

We emphasizes that, besides the choice of the reconstruction loss, the structure of the lens architecture is also important for the type of visual features removed. Its capacity and receptive field impacts the type and abstractness of visual features the lens can manipulate. We deliberately chose an architecture with skip connections to simplify learning an identity mapping that minimizes the reconstruction loss.

\begin{figure}[t]
    \centering
    \hfil
    \includegraphics[height=1.5in,trim={0 0 0.1in 0},clip]{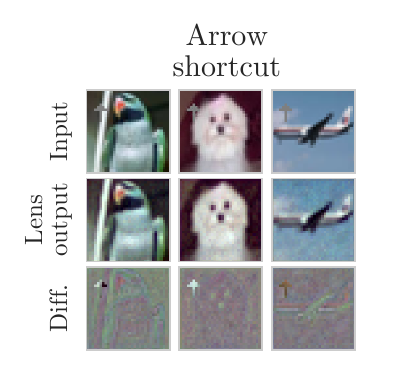}
    \includegraphics[height=1.5in,trim={0.2in 0 0 0},clip]{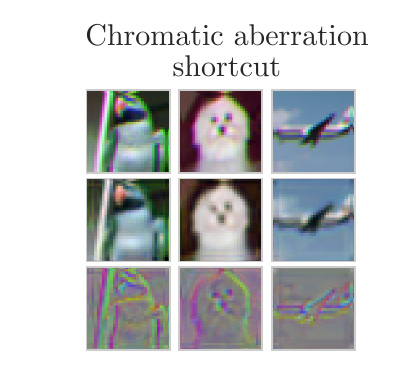}
    \hfil
    \hfil
    \vspace{3mm}
    \small
    \begin{tabular}{lll}
    \toprule
    Shortcut & \multicolumn{2}{c}{Method} \\
    \cmidrule[0.5pt]{2-3}
      & Baseline & Lens \\
    \midrule
    Arrow & $17.8 \pm 0.53$ & $76.8 \pm 0.40$ ($\mathbf{+59.01}$) \\
    Chromatic & $19.3 \pm 2.42$ & $73.5 \pm 1.23$ ($\mathbf{+54.21}$) \\
    \bottomrule
    \end{tabular}
    \vspace{3mm}
    \caption{\textbf{Top:} Example images from \cifar{} with two different synthetic shortcuts for the \rotation{} task (best viewed on screen). The \emph{Arrow} shortcut adds directional information to a small region of the image (note the arrow in the top left of the image). The \emph{Chromatic aberration} shortcut shifts the color channels, adding directional information globally. The lens learns to remove the shortcuts.
    \textbf{Bottom:} Downstream testing accuracy (in \%) of a logistic regression model trained for \cifar{} classification on the frozen representations. Accuracy on clean data is $81.8 \pm 0.30$.
    }
    \label{fig:cifar_synthetic_shortcuts}
    \vspace{-5mm}
\end{figure}

\paragraph{Comparison with standard adversarial training.} One might wonder why we do not just use standard adversarial training methods such as the FGSM \cite{goodfellow2014explaining} instead of the lens.\footnote{Iterative techniques such as projected gradient descent \cite{madry2017towards} are too expensive for our purposes.} Besides outperforming the FGSM empirically (see Section~\ref{sec:experiments}), the lens has several other advantages. First, standard adversarial training requires a loss to generate a perturbation whereas our lens can be used independently of the pretext task used during training. Hence, the lens can be deployed downstream even when the pretext task loss and/or the feature extraction network are unavailable. Second, training the lens is of similar complexity as one-step adversarial training techniques, but deploying it downstream is much cheaper as its application only requires a single forward pass through the shallow lens network (and not a forward and backward pass through the representation network as in adversarial training). The same advantages apply for visualization of the lens action. Finally, we believe that lens learns to exploit similarities and structure that is shared between shortcuts as it accumulates signal from all training examples.
\vspace{15mm}
\section{Experiments} \label{sec:experiments}

\subsection{Proof of concept: Removing synthetic shortcuts}

Our approach rests on the hypothesis that features which are easy to exploit for solving self-supervised tasks are also easy to learn for the adversarial lens to remove. We initially test this hypothesis in a controlled experimental setup, by adding synthetic shortcuts to the data. 

We use the \cifar{} dataset and the \rotation{} pretext task \cite{gidaris2018unsupervised}. In this task, each input image is fed to the network in four copies, rotated by 0\degree{}, 90\degree{}, 180\degree{} and 270\degree{}, respectively. The network is trained to solve the four-way classification task of predicting the correct orientation of each image. The task is motivated by the hypothesis that the network needs to learn high-level object representations to predict image rotations.

\begin{table*}[t]\small
    \caption{Evaluation of representations from models trained on \imagenet{} with different self-supervised pretext tasks. The scores are accuracies (in \%) of a logistic regression model trained on representations obtained from the frozen models. Mean $\pm$ s.e.m over three random initializations. Values in bold are better than the next-best method at a significance level of 0.05. Training images are preprocessed as suggested by the respective original works.}
    \label{table:main_results_table}
    \vspace{0.3cm}
    \centering
    \begin{tabular}{llllll}
    \toprule
    Dataset & Method & \multicolumn{4}{c}{Pretext task} \\
    \cmidrule[0.5pt]{3-6}
              & &                            Rotation &                            Exemplar &                     Rel. patch loc. &                              Jigsaw \\
    \midrule
    ImageNet & Baseline &                     $46.6 \pm 0.02$ &                     $43.7 \pm 0.25$ &                     $40.2 \pm 0.13$ &                     $37.2 \pm 0.06$ \\
              & FGSM &         $48.1 \pm 0.04$ $({+1.45})$ &         $44.6 \pm 0.27$ $({+0.98})$ &         $41.9 \pm 0.10$ $({+1.71})$ &         $39.3 \pm 0.38$ $({+2.10})$ \\
              & Lens &  $48.6 \pm 0.04$ $(\mathbf{+1.95})$ &  $46.1 \pm 0.04$ $(\mathbf{+2.40})$ &         $42.1 \pm 0.05$ $({+1.83})$ &  $40.9 \pm 0.11$ $(\mathbf{+3.69})$ \\
    
    \midrule
    Places205 & Baseline &                     $39.2 \pm 0.07$ &                     $41.2 \pm 0.21$ &                     $41.1 \pm 0.12$ &                     $39.0 \pm 0.23$ \\
              & FGSM &         $39.8 \pm 0.14$ $({+0.65})$ &         $39.9 \pm 0.42$ $({-1.38})$ &         $41.6 \pm 0.17$ $({+0.44})$ &         $38.8 \pm 0.24$ $({-0.14})$ \\
              & Lens &  $40.6 \pm 0.14$ $(\mathbf{+1.38})$ &  $42.4 \pm 0.22$ $(\mathbf{+1.20})$ &  $42.4 \pm 0.08$ $(\mathbf{+1.26})$ &  $40.9 \pm 0.04$ $(\mathbf{+1.99})$ \\
    \bottomrule
    \end{tabular}
\end{table*}
To test the lens, we add shortcuts to the input images that are designed to contain directional information and allow solving the \rotation{} task without the need to learn object-level features (Figure~\ref{fig:cifar_synthetic_shortcuts}, top). Representations learned by the baseline network (without lens) from data with synthetic shortcuts perform poorly downstream (Figure~\ref{fig:cifar_synthetic_shortcuts}, bottom). In contrast, feature extractors learned with the lens perform dramatically better. The lens output images reveal that the lens learns to remove the synthetic shortcuts. These results confirm our hypothesis that an adversarially trained lens can learn to remove shortcut features from the data.

\subsection{Large-scale evaluation of the lens performance}

To test the value of the lens as a general framework for improving self-supervised representation learning, we next evaluate the method on two large-scale datasets and four common pretext tasks for which open-source reference implementations are available \cite{kolesnikov2019revisiting}\footnote{https://github.com/google/revisiting-self-supervised}. 

\subsubsection{Pretext tasks}
In addition to the \rotation{} task described above, we use the following pretext tasks:

\textbf{Exemplar} \cite{dosovitskiy2014discriminative}: In the \exemplar{} task, eight copies are created of each image and augmented separately using random translation, scaling, brightness and saturation, including conversion to grayscale with probability 0.66. The network is trained using a triplet loss \cite{schroff2015facenet} to minimize the embedding distance between copies of the same image, while maximizing their distance to the other images in the batch.

\textbf{Relative patch location} \cite{noroozi2016unsupervised}: Here, the task is to predict the relative location of an image patch in the 8-connected neighborhood of a reference patch from the same image (e.g. ``below'', ``upper left'' etc.). 

\textbf{Jigsaw} \cite{doersch2015unsupervised}: For the \jigsaw{} task, the image is divided into a three-by-three grid of patches. The patch order is randomly permuted and the patches are fed through the network. The representations produced by the network are then passed through a two-layer perceptron, which predicts the patch permutation.

For the patch-based tasks, we obtain representations for evaluation by averaging the representations of nine non-augmented patches created by dividing the input image into a regular three-by-three grid \cite{kolesnikov2019revisiting}.

\subsubsection{Pretraining datasets and preprocessing}
Self-supervised training is performed on \imagenet{}, which contains 1.3 million images, each belonging to one of 1000 object categories. 
Unless stated otherwise, we use the same preprocessing operations and batch size as \citet{kolesnikov2019revisiting} for the respective tasks. To mitigate distribution shift between raw and lens-processed images, we feed both the batch of lens-processed and the raw images to the feature extraction network (\citet{kurakin2016adversarial} similarly feed processed and raw images for adversarial training). This is done for all tasks except for \exemplar{}, for which memory constraints did not allow inclusion of unprocessed images. Training was performed on 128 TPU v3 cores for \rotation{} and \exemplar{} and 32 TPU v3 cores for \relpatchloc{} and \jigsaw{}.

\subsubsection{Pretraining hyperparameters}
Feature extractor and lens are trained synchronously using the Adam optimizer with $\beta_1=0.1$, $\beta_2=10^{-3}$ and $\epsilon = 10^{-7}$ for 35 epochs. The learning rate is linearly ramped up from zero to $10^{-4}$ in the first epoch, stays at $10^{-4}$ until the end of the 32\textsuperscript{nd} epoch, and is then linearly decayed to zero.

For each pretext task, we initially roughly determined the appropriate magnitude of $\reclossscale$ based on the magnitude of the pretext task loss. We then sweep over five values of $\reclossscale$, centered on the previously determined value, and report the accuracy for the best $\reclossscale$.

\subsubsection{Evaluation protocol}
\label{subsec:evaluation}
To evaluate the quality of the representations learned by the feature extractor, we follow the common \emph{linear evaluation} protocol \cite{dosovitskiy2014discriminative,doersch2015unsupervised,noroozi2016unsupervised,gidaris2018unsupervised}: We obtain image representations from the frozen feature extractor and use them to train a logistic regression model to solve multi-class image classification tasks. For networks trained with the lens, we concatenate features from the \emph{pre-logits} layer obtained with and without applying the lens to the input image (see Section~\ref{sec:shortcut-def}). To ensure a fair comparison, for baseline models (trained without lens), we concatenate two copies of the \emph{pre-logits} features to match the representation size of the lens networks. Note that the representation size determines the number of free parameters of the logistic regression model and we observed benefits in logistic regression accuracy (possibly due to a change in optimization dynamics) even though no new features are added by copying them.  The logistic regression model is optimized by stochastic gradient descent (see Appendix~\ref{app:evaluation}).

We report top-1 classification accuracy on the \imagenet{} validation set. In addition, to measure how well the learned representations transfer to unseen data, we also report downstream top-1 accuracy on the \places{} dataset. This dataset contains 2.5M images from 205 scene classes such as \emph{airfield}, \emph{kitchen}, \emph{coast} etc.

\subsubsection{Results}
\label{subsec:results}

\paragraph{Representation quality.} Table~\ref{table:main_results_table} shows evaluation results across tasks and datasets. For all tested tasks, adding the lens leads to a significant improvement over the baseline. Further, the lens outperforms adversarial training using the fast gradient sign method (FGSM; \citealp{goodfellow2014explaining}; details in Appendix~\ref{app:fgsm}). In particular, the lens outperforms the FGSM by a large margin when \imagenet{}-trained representations are transferred to the \places{}-dataset (Table~\ref{table:main_results_table}, bottom). The improved transfer performance suggests that the features modified by the lens are more general than those attacked by the FGSM.

Our results show that the benefit of automatic shortcut removal using an adversarially trained lens generalizes across pretext tasks and across datasets. Furthermore, we find that gains can be observed across a wide range of feature extractor capacities (model widths; see \supplement{}).

\begin{figure}[hb]
\begin{center}
    \vspace{3mm}
    \centerline{\includegraphics[width=2.8in]{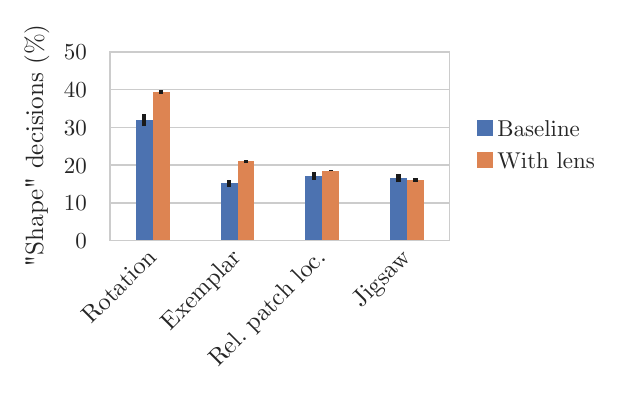}}
    \vspace{-5mm}
    \caption{Percentage of shape-based rather than texture-based decisions on the cue conflict dataset from \citet{geirhos2019texture}. Error bars show the mean$\pm$s.e.m. over three random initializations.}
    \label{fig:geirhos_evaluation}
\end{center}
\end{figure}

\paragraph{Increased semanticity of representations.}
To test whether shortcut removal leads to higher-level, more semantically meaningful representations, we evaluated networks on images with conflicting texture and shape information \cite{geirhos2019texture}. CNNs are typically biased towards using textures (i.e. low-level features) for solving classification tasks, in contrast to humans, who use shapes \cite{geirhos2019texture}. We find that shortcut removal shifts networks towards using more shape information (Figure~\ref{fig:geirhos_evaluation}). In addition, the lensed representations perform better than baseline even without concatenating unlensed representations as described above (see Table~\ref{table:results_table_without_concatenation} in the \supplement{}), although concatenation leads to further improvement. These results suggest that our method encourages networks to learn more semantic representations.

\begin{figure}[ht]
\begin{center}
    \centerline{\includegraphics[width=\columnwidth]{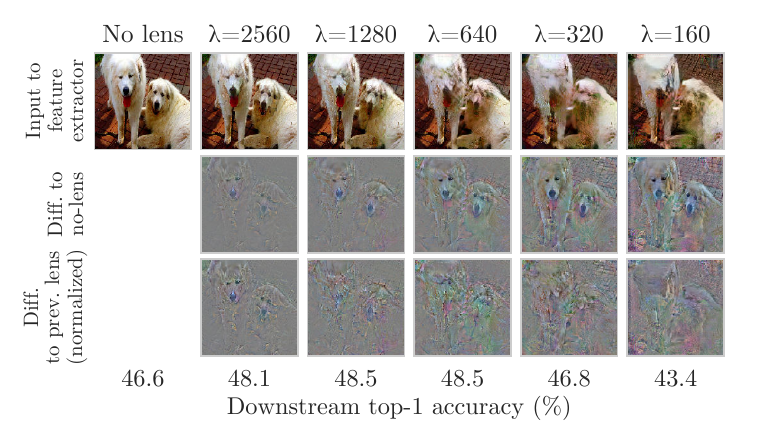}}
    \caption{Lens outputs for \rotation{} models trained with different reconstruction loss scales $\reclossscale$ (best viewed on screen). Decreasing $\reclossscale$ allows the lens to make increasingly large changes to the image and reveals the relative importance of features for the pretext task. For example, eyes and nose are affected at the highest $\reclossscale$ (2560), while the legs are affected only at lower values (320).}
    \label{fig:reconstruction_loss_scale_sweep}
\end{center}
\end{figure}

\paragraph{Visualization of removed features.} To understand what features the lens removes, we visualize the input image, processed image, and their difference. The lens network produces visually interpretable modifications, in contrast to single-step adversarial attacks such as FGSM \cite{goodfellow2014explaining}. Figure~\ref{fig:reconstruction_loss_scale_sweep} illustrates how an image is modified by lens networks trained with different values for the reconstruction loss scale $\reclossscale$ on the \rotation{} task. The progression of image changes with decreasing $\reclossscale$ reveal what features are preferentially used by the feature extraction network to solve the \rotation{} task.

\paragraph{SimCLR.} Concurrently with our work, a powerful new self-supervised approach based on contrastive learning, called \emph{SimCLR}, was published \cite{chen2020simple}. In Appendix~\ref{sec:simclr}, we describe our experience applying automatic shortcut removal to \emph{SimCLR} as a ``case study''. While shortcut removal does not improve linear evaluation performance of \emph{SimCLR} on \imagenet{}, we find that our method automatically discovers some of the hand-designed preprocessing operations used to train \emph{SimCLR} and strongly increases the semanticity of learned representations, leading to gains on a wide range of other tasks.

\subsection{Comparing lens features across pretext tasks} \label{subsec:comparison_across_tasks}

\begin{figure}[t]
    \centering
    \begin{subfigure}[t]{\columnwidth}
    \includegraphics[width=1.0\linewidth]{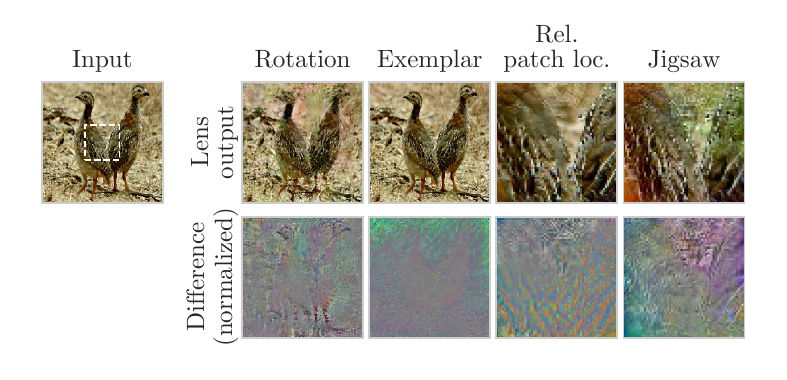}
    \end{subfigure}
    \vskip -0.1in
    \begin{subfigure}[t]{\columnwidth}
        \includegraphics[width=1.0\linewidth,trim={0 0 0 0.32in},clip]{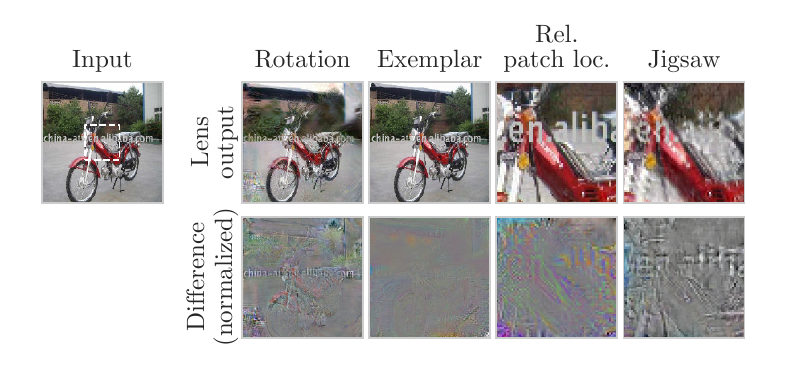}
    \end{subfigure}
    \vskip -0.1in
    \begin{subfigure}[t]{\columnwidth}
        \includegraphics[width=1.0\linewidth,trim={0 0 0 0.32in},clip]{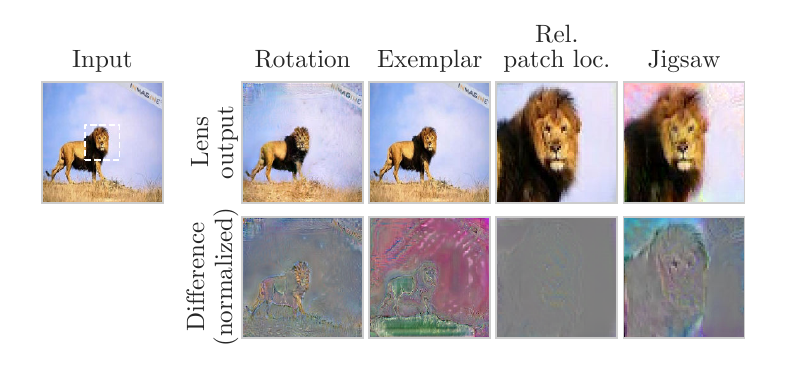}
    \end{subfigure}
    \vskip -0.1in
    \begin{subfigure}[t]{\columnwidth}
        \includegraphics[width=1.0\linewidth,trim={0 0.65in 0 0},clip]{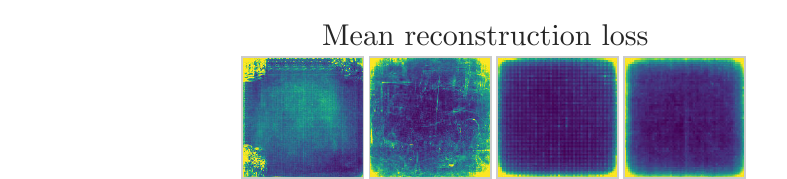}
    \end{subfigure}

    \caption{\textbf{Top:} Three example images from \imagenet{}, processed by lenses trained on different pretext tasks (best viewed on screen). The dashed square on the input image shows the region used for the patch-based tasks (\relpatchloc{} and \jigsaw{}). 
    \textbf{Bottom:} Mean rec. loss across 1280 images randomly chosen from the test set. For display, values were clipped at the 95\textsuperscript{th} percentile.
    }
    \label{fig:comparison_across_tasks}
\end{figure}

\paragraph{Qualitative comparison of shortcut features.} The trained lens represents a new tool for visualizing and comparing the features learned by different pretext tasks. Inspection of lens outputs confirms existing heuristics and provides new insights (Figure~\ref{fig:comparison_across_tasks}), as discussed in the following.

The features removed by the lens are the most semantically meaningful for the \rotation{} task, potentially explaining why its representations outperform the other tasks. The lens removes features such as head and feet of animals, and is generally biased towards the image center (see mean reconstruction loss images, Figure~\ref{fig:comparison_across_tasks}, bottom). Text and watermarks provide a strong orientation signal and are also concealed, which is reflected by high mean reconstruction loss values in the corners of the image, where logos are often found. These results confirm expectations and support the validity of lens outputs as an interpretability tool.

For the \exemplar{} task, the lens introduces full-field color changes, suggesting that this task is easily solved by matching images based on their average/dominant color. In contrast to \rotation{}, the mean reconstruction loss is biased away from the image center.

\begin{figure}[tb]
\begin{center}
    \vspace{5mm}
    \centerline{\includegraphics[width=\columnwidth]{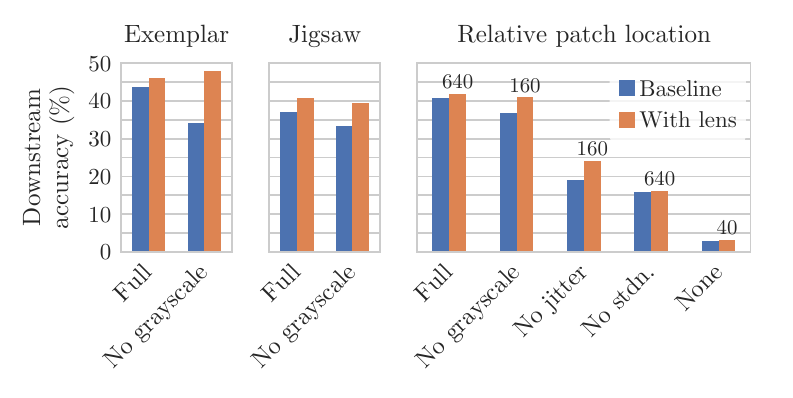}}
    \vspace{-3mm}
    \caption{Downstream accuracy on \imagenet{} for tasks that use conversion to grayscale for augmentation in their reference implementations. The lens always outperforms this manual augmentation. For \relpatchloc{}, we also ablated all other manual augmentations. The full augmentation pipeline for \relpatchloc{} involves: (1) conversion to grayscale with probability 0.66, (2) independent jittering of patch locations by up to 21 pixels, and (3) independent color standardization of patches. The number above the bars indicates the optimal reconstruction loss scale $\reclossscale{}$ based on a sweep over $\{40, 80, 160, 320, 640\}$.}
    \label{fig:augmentation_ablation_combined}
    \vspace{-4mm}
\end{center}
\end{figure} 

\begin{figure}[ht]
\begin{center}
    \vspace{1.5mm}
    \centerline{\includegraphics[width=\columnwidth]{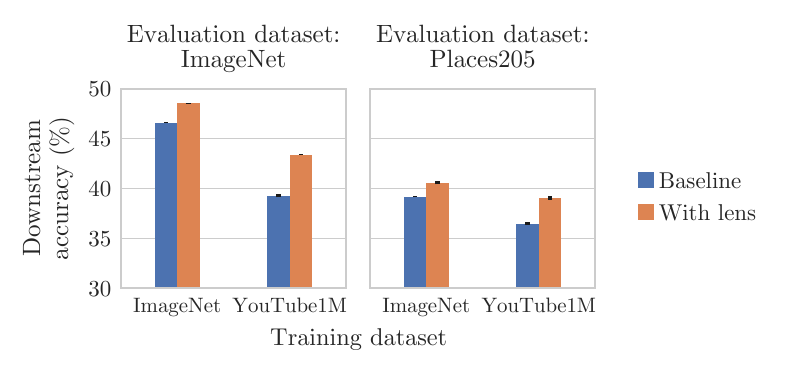}}
    \caption{Downstream \imagenet{} classification accuracy for models trained on \imagenet{} or \youtube{}. The lens recovers much of the accuracy lost when training on the less curated \youtube{} dataset. Error bars: mean$\pm$s.e.m. over three random initializations.}
    \label{fig:youtube_accuracy}
\end{center}
\end{figure}

For patch-based tasks such as \relpatchloc{} and \jigsaw{}, much effort has gone into identifying trivial solutions and designing augmentation schemes to reduce their effect. These shortcuts can be hard to identify. For example, in the paper introducing the \relpatchloc{} task, \citet{doersch2015unsupervised} express their surprise at finding that convolutional neural networks learn to exploit chromatic aberrations to solve the task. To mitigate the shortcut, they drop color channels at random during training. Similar augmentations are proposed for the \jigsaw{} task \cite{noroozi2016unsupervised}. More recently, refined color augmentation heuristics such as \emph{chroma blurring} have been developed for patch-based pretext tasks \cite{mundhenk2018improvements}.

\looseness-1Our approach learns similar augmentations automatically, and provides additional insights. Specifically, the lens output (Figure~\ref{fig:comparison_across_tasks}) suggests that that color augmentations such as chroma blurring improve \relpatchloc{} and \jigsaw{} for different reasons: For \relpatchloc{}, color fringes at high-contrast edges (as caused by chromatic aberrations) are the most prominent visual feature modified by the lens. In contrast, the lens effect for \jigsaw{} involves diffuse color changes towards the image borders, suggesting that color matching across patch borders is a major shortcut. The difference images also suggest that \jigsaw{}, but not \relpatchloc{}, additionally benefits from luminance blurring, because luminance edges are prominent in the Jigsaw difference images. 

\paragraph{Ablation of preprocessing operations.} Quantitatively, we find that random color dropping becomes less relevant when using the lens (Figure~\ref{fig:augmentation_ablation_combined}): In all cases, the lens can at least compensate for the missing color dropping operation, and for \exemplar{} even performs better without color dropping than with it. For \relpatchloc{}, we additionally perform an ablation analysis of the whole augmentation pipeline (Figure~\ref{fig:augmentation_ablation_combined}). We find that the lens can replace color dropping and, to some degree, random patch jitter. However, color standardization remains important, likely because full-field color shifts are expensive under the $\ell_2$ reconstruction loss.

\begin{figure*}[t]
\setlength{\fboxsep}{0.05in}
\setlength{\fboxrule}{0.25pt}
    \fbox{\includegraphics[height=2.1in,trim={0.09in 0.15in 0.15in 0.1in},clip]{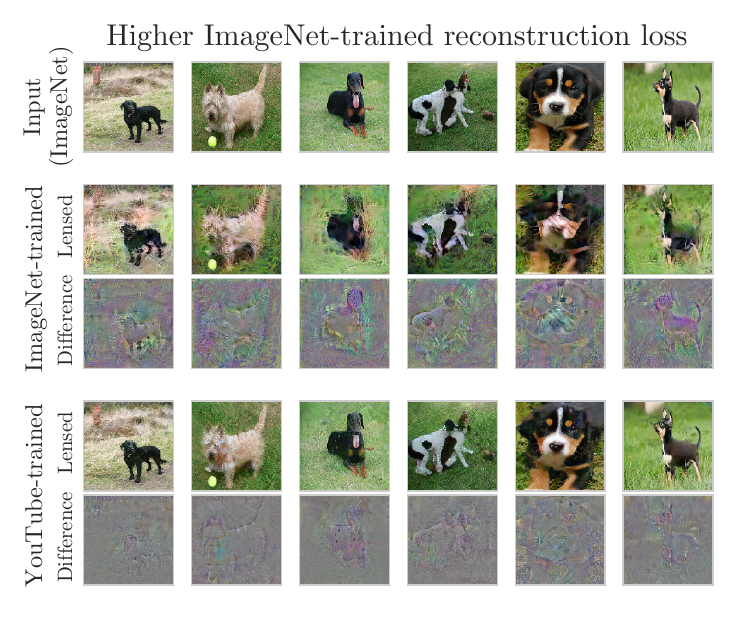}}
    \hfill
    \fbox{\includegraphics[height=2.1in,trim={0.33in 0.15in 0.15in 0.1in},clip]{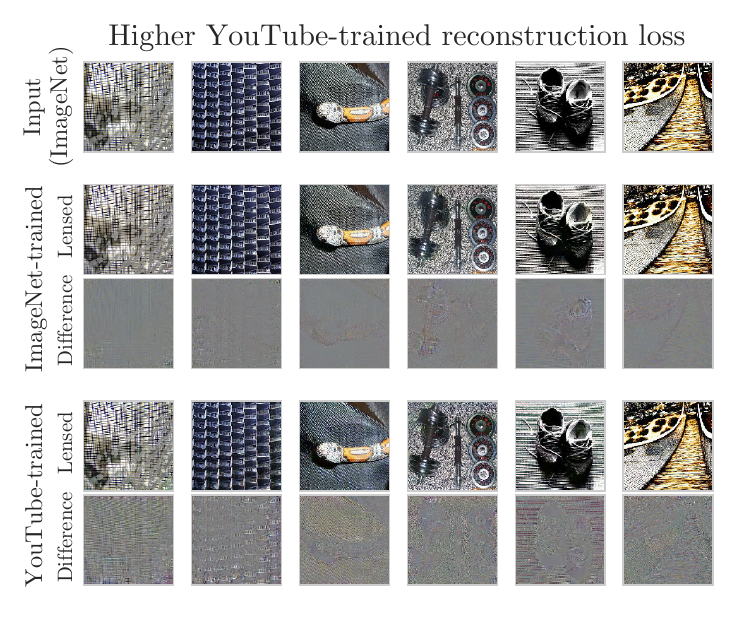}}
    \hfill
    \hfill
    \fbox{\includegraphics[height=2.1in,trim={0.33in 0.15in 0.15in 0.1in},clip]{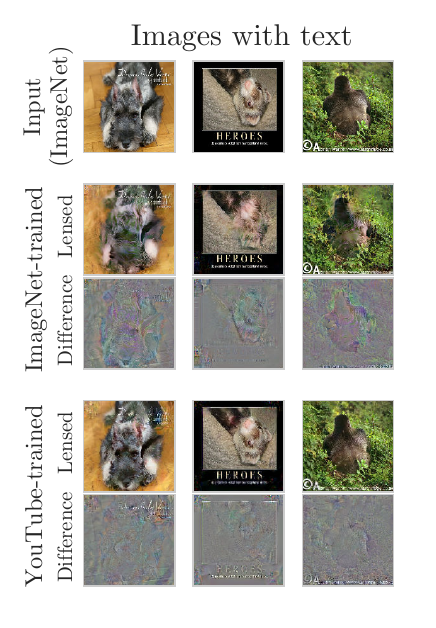}}
    \addtocounter{figure}{1}  
    \caption{Comparison of lens outputs for models trained on \imagenet{} and \youtube{} (best viewed on screen). Images are ordered based on the difference in reconstruction loss between the \imagenet{}-trained and \youtube{}-trained models. The left block shows the top six images (i.e. higher reconstruction loss when trained on \imagenet{}); the right block shows the bottom six images (i.e. higher reconstruction loss when trained on \youtube{}). The modifications made by the lens can thus be used to identify dataset bias. The three images on the far right were hand-selected to contain text and illustrate that \youtube{}-trained models are more sensitive to this shortcut.}
    \label{fig:youtube_examples}
    \vspace{4mm}
\end{figure*}

 \paragraph{Similarity of shortcuts across tasks.} The lens output allows for a quantitative evaluation of shortcut similarity. Specifically, the correlations between the per-image reconstruction loss across tasks suggest that patch-based \begin{wrapfigure}[15]{r}{0.4\columnwidth}
    \centering
    \includegraphics[width=0.4\columnwidth,trim={0.1in 0.1in 0.1in 0.1in},clip]{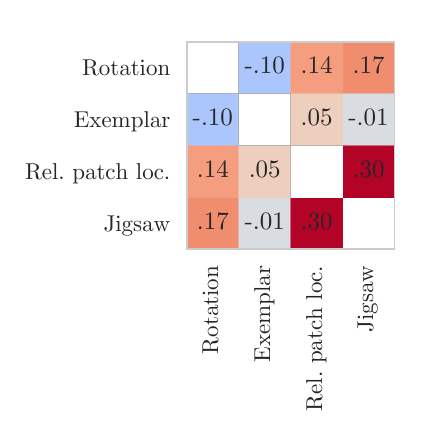}
    \addtocounter{figure}{-2} 
    \vspace{-6mm}
    \caption{Pearson correlation of the per-image reconstruction loss between pretext tasks (\imagenet{}).}
    \label{fig:task_correlation}
    \vspace{-2mm}
\end{wrapfigure} tasks are vulnerable to similar shortcuts, whereas \rotation{} and \exemplar{} show anti-correlated reconstruction losses (Figure~\ref{fig:task_correlation}). This suggests that the \rotation{} and \exemplar{}, but less so \jigsaw{} and \relpatchloc{}, may combine synergistically in multi-task learning. Empirical gains through training multiple pretext task jointly have been observed by \citet{doersch2017multi}.

\subsection{Comparing lens features across datasets}

Apart from the choice of pretext task, the pretraining dataset is another major factor influencing the representations learned by self-supervision. Since self-supervision does not require human-annotated labels, it opens new possibilities for mining large datasets from abundant unlabeled sources such as videos from the web. However, such data sources may contain more shortcuts (and less informative images) than highly curated pure image data sets, e.g. logos of TV stations, text from credits, black frames, etc. As a result, certain pretext tasks suffer a significant performance drop when applied to frames mined from videos \cite{tschannen2019self}. We therefore tested the effectiveness of the lens in self-supervised learning on video frames, using the \rotation{} task. For training, we used 1 million randomly sampled frames from the YouTube-8M dataset \cite{abu2016youtube} as in \cite{tschannen2019self} (referred to as \youtube{}). As expected, performance drops compared to training on \imagenet{}, but the performance reduction can be compensated at least partially by the lens (Figure~\ref{fig:youtube_accuracy}). The lens recovers about half of the gap to \imagenet{}-training when evaluating on \imagenet{} classification downstream. In the transfer setting, when evaluating on \places{}, the lens performs even better, such that the \youtube{}-trained model with lens performs similarly to the \imagenet{}-trained baseline.

Inspecting the lens outputs for models trained on \imagenet{} and \youtube{} allows us to compare the shortcut features across these datasets. In Figure~\ref{fig:youtube_examples}, we show the images with the largest difference in reconstruction loss when trained on \imagenet{} or \youtube{}. The images strikingly align with the biases of the respective datasets: For \imagenet{}, all of the top images contain dogs, which are overrepresented in \imagenet{}. For \youtube{}, the top images predominantly show high-contrast edges oriented along the cardinal directions. We speculate that this is because images with black bars (for aspect ratio conversion) may be common in this video-derived dataset. These results suggest that in biased datasets, the overrepresented features are easier to learn and thus become shortcuts relative to underrepresented features. The lens learns to target these overrepresented classes and thereby counteracts datasets biases. 

We also found that the \youtube{}-trained lens is more sensitive to overlaid text than the \imagenet{}-trained lens (Figure~\ref{fig:youtube_examples}, right). Overlaid text is common in \youtube{} (e.g. video credits), but less so in \imagenet{}.

Together, our quantitative and qualitative results show that the lens can be used to identify and remove pretext task-specific biases and shortcut features from datasets. The lens is therefore a promising tool for exploiting large non-curated data sources.

\section{Conclusion}

We proposed a method to improve self-supervised visual representations by learning to remove shortcut features from the data with an adversarially trained lens network. Training a feature extractor on data from which shortcuts have been removed forces it to learn higher-level features that transfer and generalize better, as shown in our experiments. By combining features obtained with and without shortcut removal into a single image representation, we ensure that our approach improves representation quality even if the removed features are relevant for the downstream task.

Apart from improved representations, our approach allows us to visualize, quantify and compare the features learned by self-supervision. We confirm that our approach detects and mitigates shortcuts observed in prior work and also sheds light on issues that were less known.

Future research should explore design choices such as the lens architecture and image reconstruction loss. Furthermore, it would be great to see whether and how the proposed technique can be applied to improve and/or visualize supervised learning algorithms.

\section*{Acknowledgements}
We thank Xiaohua Zhai for help with the self-supervised learning code base. We also thank Sylvain Gelly and the Google Brain team in Zurich for helpful discussions

\clearpage

\bibliography{references}

\begin{thebibliography}{42}
\providecommand{\natexlab}[1]{#1}
\providecommand{\url}[1]{\texttt{#1}}
\expandafter\ifx\csname urlstyle\endcsname\relax
  \providecommand{\doi}[1]{doi: #1}\else
  \providecommand{\doi}{doi: \begingroup \urlstyle{rm}\Url}\fi

\bibitem[Abu-El-Haija et~al.(2016)Abu-El-Haija, Kothari, Lee, Natsev, Toderici,
  Varadarajan, and Vijayanarasimhan]{abu2016youtube}
Abu-El-Haija, S., Kothari, N., Lee, J., Natsev, P., Toderici, G., Varadarajan,
  B., and Vijayanarasimhan, S.
\newblock Youtube-8m: A large-scale video classification benchmark.
\newblock \emph{arXiv:1609.08675}, 2016.

\bibitem[Akhtar \& Mian(2018)Akhtar and Mian]{akhtar2018threat}
Akhtar, N. and Mian, A.
\newblock Threat of adversarial attacks on deep learning in computer vision: A
  survey.
\newblock \emph{IEEE Access}, 6:\penalty0 14410--14430, 2018.

\bibitem[Bachman et~al.(2019)Bachman, Hjelm, and
  Buchwalter]{bachman2019learning}
Bachman, P., Hjelm, R.~D., and Buchwalter, W.
\newblock Learning representations by maximizing mutual information across
  views.
\newblock In \emph{NeurIPS}, 2019.

\bibitem[Caron et~al.(2018)Caron, Bojanowski, Joulin, and Douze]{caron2018deep}
Caron, M., Bojanowski, P., Joulin, A., and Douze, M.
\newblock Deep clustering for unsupervised learning of visual features.
\newblock \emph{Proc. ECCV}, 2018.

\bibitem[Chen et~al.(2020)Chen, Kornblith, Norouzi, and Hinton]{chen2020simple}
Chen, T., Kornblith, S., Norouzi, M., and Hinton, G.
\newblock A simple framework for contrastive learning of visual
  representations.
\newblock \emph{Proc. ICML}, 2020.

\bibitem[Doersch \& Zisserman(2017)Doersch and Zisserman]{doersch2017multi}
Doersch, C. and Zisserman, A.
\newblock Multi-task self-supervised visual learning.
\newblock In \emph{Proc. ICCV}, 2017.

\bibitem[Doersch et~al.(2015)Doersch, Gupta, and
  Efros]{doersch2015unsupervised}
Doersch, C., Gupta, A., and Efros, A.~A.
\newblock Unsupervised visual representation learning by context prediction.
\newblock In \emph{Proc. ICCV}, pp.\  1422--1430, 2015.

\bibitem[Dosovitskiy et~al.(2014)Dosovitskiy, Springenberg, Riedmiller, and
  Brox]{dosovitskiy2014discriminative}
Dosovitskiy, A., Springenberg, J.~T., Riedmiller, M., and Brox, T.
\newblock Discriminative unsupervised feature learning with convolutional
  neural networks.
\newblock In \emph{NeurIPS}, pp.\  766--774, 2014.

\bibitem[Feng et~al.(2019)Feng, Xu, and Tao]{feng2019self}
Feng, Z., Xu, C., and Tao, D.
\newblock Self-supervised representation learning by rotation feature
  decoupling.
\newblock In \emph{Proc. CVPR}, pp.\  10364--10374, 2019.

\bibitem[Geirhos et~al.(2019)Geirhos, Rubisch, Michaelis, Bethge, Wichmann, and
  Brendel]{geirhos2019texture}
Geirhos, R., Rubisch, P., Michaelis, C., Bethge, M., Wichmann, F.~A., and
  Brendel, W.
\newblock {ImageNet-trained} {CNNs} are biased towards texture; increasing
  shape bias improves accuracy and robustness.
\newblock \emph{Proc. ICLR}, 2019.

\bibitem[Gidaris et~al.(2018)Gidaris, Singh, and
  Komodakis]{gidaris2018unsupervised}
Gidaris, S., Singh, P., and Komodakis, N.
\newblock Unsupervised representation learning by predicting image rotations.
\newblock \emph{Proc. ICLR}, 2018.

\bibitem[Goodfellow et~al.(2015)Goodfellow, Shlens, and
  Szegedy]{goodfellow2014explaining}
Goodfellow, I.~J., Shlens, J., and Szegedy, C.
\newblock Explaining and harnessing adversarial examples.
\newblock \emph{Proc. ICLR}, 2015.

\bibitem[Goyal et~al.(2019)Goyal, Mahajan, Gupta, and Misra]{goyal2019scaling}
Goyal, P., Mahajan, D., Gupta, A., and Misra, I.
\newblock Scaling and benchmarking self-supervised visual representation
  learning.
\newblock In \emph{Proc. ICCV}, pp.\  6391--6400, 2019.

\bibitem[He et~al.(2016)He, Zhang, Ren, and Sun]{he2016identity}
He, K., Zhang, X., Ren, S., and Sun, J.
\newblock Identity mappings in deep residual networks.
\newblock In \emph{Proc. ECCV}. Springer, 2016.

\bibitem[He et~al.(2020)He, Fan, Wu, Xie, and Girshick]{he2019momentum}
He, K., Fan, H., Wu, Y., Xie, S., and Girshick, R.
\newblock Momentum contrast for unsupervised visual representation learning.
\newblock In \emph{Proc. CVPR}, 2020.

\bibitem[H{\'e}naff et~al.(2019)H{\'e}naff, Razavi, Doersch, Eslami, and
  Oord]{henaff2019data}
H{\'e}naff, O.~J., Razavi, A., Doersch, C., Eslami, S., and Oord, A. v.~d.
\newblock Data-efficient image recognition with contrastive predictive coding.
\newblock \emph{arXiv:1905.09272}, 2019.

\bibitem[Hjelm et~al.(2019)Hjelm, Fedorov, Lavoie-Marchildon, Grewal, Bachman,
  Trischler, and Bengio]{hjelm2018learning}
Hjelm, R.~D., Fedorov, A., Lavoie-Marchildon, S., Grewal, K., Bachman, P.,
  Trischler, A., and Bengio, Y.
\newblock Learning deep representations by mutual information estimation and
  maximization.
\newblock In \emph{Proc. ICLR}, 2019.

\bibitem[Jenni \& Favaro(2018)Jenni and Favaro]{jenni2018artifacts}
Jenni, S. and Favaro, P.
\newblock {Self-Supervised} feature learning by learning to spot artifacts.
\newblock \emph{Proc. CVPR}, 2018.

\bibitem[Kolesnikov et~al.(2019)Kolesnikov, Zhai, and
  Beyer]{kolesnikov2019revisiting}
Kolesnikov, A., Zhai, X., and Beyer, L.
\newblock Revisiting self-supervised visual representation learning.
\newblock In \emph{Proc. CVPR}, pp.\  1920--1929, 2019.

\bibitem[Kurakin et~al.(2016)Kurakin, Goodfellow, and
  Bengio]{kurakin2016adversarial}
Kurakin, A., Goodfellow, I., and Bengio, S.
\newblock Adversarial machine learning at scale.
\newblock In \emph{Proc. ICLR}, 2016.

\bibitem[Lee et~al.(2017)Lee, Huang, Singh, and Yang]{lee2017unsupervised}
Lee, H.-Y., Huang, J.-B., Singh, M., and Yang, M.-H.
\newblock Unsupervised representation learning by sorting sequences.
\newblock In \emph{Proc. ICCV}, pp.\  667--676, 2017.

\bibitem[Madry et~al.(2018)Madry, Makelov, Schmidt, Tsipras, and
  Vladu]{madry2017towards}
Madry, A., Makelov, A., Schmidt, L., Tsipras, D., and Vladu, A.
\newblock Towards deep learning models resistant to adversarial attacks.
\newblock In \emph{Proc. ICLR}, 2018.

\bibitem[Misra \& van~der Maaten(2019)Misra and van~der Maaten]{misra2019self}
Misra, I. and van~der Maaten, L.
\newblock Self-supervised learning of pretext-invariant representations.
\newblock \emph{arXiv:1912.01991}, 2019.

\bibitem[Miyato et~al.(2018)Miyato, Maeda, Koyama, and
  Ishii]{miyato2018virtual}
Miyato, T., Maeda, S.-i., Koyama, M., and Ishii, S.
\newblock Virtual adversarial training: a regularization method for supervised
  and semi-supervised learning.
\newblock \emph{IEEE Transactions on Pattern Analysis and Machine
  Intelligence}, 41\penalty0 (8):\penalty0 1979--1993, 2018.

\bibitem[Mundhenk et~al.(2018)Mundhenk, Ho, and Chen]{mundhenk2018improvements}
Mundhenk, T.~N., Ho, D., and Chen, B.~Y.
\newblock Improvements to context based self-supervised learning.
\newblock In \emph{Proc. CVPR}, 2018.

\bibitem[Noroozi \& Favaro(2016)Noroozi and Favaro]{noroozi2016unsupervised}
Noroozi, M. and Favaro, P.
\newblock Unsupervised learning of visual representations by solving jigsaw
  puzzles.
\newblock In \emph{Proc. ECCV}, pp.\  69--84, 2016.

\bibitem[Noroozi et~al.(2017)Noroozi, Pirsiavash, and
  Favaro]{noroozi2017representation}
Noroozi, M., Pirsiavash, H., and Favaro, P.
\newblock Representation learning by learning to count.
\newblock In \emph{Proc. ICCV}, 2017.

\bibitem[Oord et~al.(2018)Oord, Li, and Vinyals]{oord2018representation}
Oord, A. v.~d., Li, Y., and Vinyals, O.
\newblock Representation learning with contrastive predictive coding.
\newblock \emph{arXiv:1807.03748}, 2018.

\bibitem[Ronneberger et~al.(2015)Ronneberger, Fischer, and
  Brox]{ronneberger2015unet}
Ronneberger, O., Fischer, P., and Brox, T.
\newblock U-net: Convolutional networks for biomedical image segmentation.
\newblock \emph{Med. Image Comput. Comput. Assist. Interv.}, 2015.

\bibitem[Sajjadi et~al.(2018)Sajjadi, Parascandolo, Mehrjou, and
  Sch{\"o}lkopf]{sajjadi2018tempered}
Sajjadi, M.~S., Parascandolo, G., Mehrjou, A., and Sch{\"o}lkopf, B.
\newblock Tempered adversarial networks.
\newblock In \emph{Proc. ICML}, pp.\  4451--4459, 2018.

\bibitem[Salimans et~al.(2016)Salimans, Goodfellow, Zaremba, Cheung, Radford,
  and Chen]{salimans2016improved}
Salimans, T., Goodfellow, I., Zaremba, W., Cheung, V., Radford, A., and Chen,
  X.
\newblock Improved techniques for training gans.
\newblock In \emph{NeurIPS}, 2016.

\bibitem[Santurkar et~al.(2019)Santurkar, Ilyas, Tsipras, Engstrom, Tran, and
  Madry]{santurkar2019image}
Santurkar, S., Ilyas, A., Tsipras, D., Engstrom, L., Tran, B., and Madry, A.
\newblock Image synthesis with a single (robust) classifier.
\newblock In \emph{NeurIPS}, pp.\  1260--1271, 2019.

\bibitem[Schroff et~al.(2015)Schroff, Kalenichenko, and
  Philbin]{schroff2015facenet}
Schroff, F., Kalenichenko, D., and Philbin, J.
\newblock Facenet: A unified embedding for face recognition and clustering.
\newblock In \emph{Proc. CVPR}, 2015.

\bibitem[Szegedy et~al.(2014)Szegedy, Zaremba, Sutskever, Bruna, Erhan,
  Goodfellow, and Fergus]{szegedy2013intriguing}
Szegedy, C., Zaremba, W., Sutskever, I., Bruna, J., Erhan, D., Goodfellow, I.,
  and Fergus, R.
\newblock Intriguing properties of neural networks.
\newblock In \emph{Proc. ICLR}, 2014.

\bibitem[Tian et~al.(2019)Tian, Krishnan, and Isola]{tian2019contrastive}
Tian, Y., Krishnan, D., and Isola, P.
\newblock Contrastive multiview coding.
\newblock \emph{arXiv:1906.05849}, 2019.

\bibitem[Tschannen et~al.(2020{\natexlab{a}})Tschannen, Djolonga, Ritter,
  Mahendran, Houlsby, Gelly, and Lucic]{tschannen2019self}
Tschannen, M., Djolonga, J., Ritter, M., Mahendran, A., Houlsby, N., Gelly, S.,
  and Lucic, M.
\newblock Self-supervised learning of video-induced visual invariances.
\newblock In \emph{Proc. CVPR}, 2020{\natexlab{a}}.

\bibitem[Tschannen et~al.(2020{\natexlab{b}})Tschannen, Djolonga, Rubenstein,
  Gelly, and Lucic]{tschannen2020mutual}
Tschannen, M., Djolonga, J., Rubenstein, P.~K., Gelly, S., and Lucic, M.
\newblock On mutual information maximization for representation learning.
\newblock In \emph{Proc. ICLR}, 2020{\natexlab{b}}.

\bibitem[Wei et~al.(2018)Wei, Lim, Zisserman, and Freeman]{wei2018learning}
Wei, D., Lim, J.~J., Zisserman, A., and Freeman, W.~T.
\newblock Learning and using the arrow of time.
\newblock In \emph{Proc. CVPR}, pp.\  8052--8060, 2018.

\bibitem[Xie et~al.(2020)Xie, Tan, Gong, Wang, Yuille, and
  Le]{xie2019adversarial}
Xie, C., Tan, M., Gong, B., Wang, J., Yuille, A., and Le, Q.~V.
\newblock Adversarial examples improve image recognition.
\newblock In \emph{Proc. CVPR}, 2020.

\bibitem[Zhai et~al.(2019)Zhai, Puigcerver, Kolesnikov, Ruyssen, Riquelme,
  Lucic, Djolonga, Pinto, Neumann, Dosovitskiy, Beyer, Bachem, Tschannen,
  Michalski, Bousquet, Gelly, and Houlsby]{zhai2019vtab}
Zhai, X., Puigcerver, J., Kolesnikov, A., Ruyssen, P., Riquelme, C., Lucic, M.,
  Djolonga, J., Pinto, A.~S., Neumann, M., Dosovitskiy, A., Beyer, L., Bachem,
  O., Tschannen, M., Michalski, M., Bousquet, O., Gelly, S., and Houlsby, N.
\newblock A large-scale study of representation learning with the visual task
  adaptation benchmark.
\newblock \emph{arXiv:1910.04867}, 2019.

\bibitem[Zhang et~al.(2016)Zhang, Isola, and Efros]{zhang2016colorful}
Zhang, R., Isola, P., and Efros, A.~A.
\newblock Colorful image colorization.
\newblock In \emph{Proc. ECCV}, pp.\  649--666, 2016.

\bibitem[Zhang et~al.(2018)Zhang, Isola, Efros, Shechtman, and
  Wang]{zhang2018unreasonable}
Zhang, R., Isola, P., Efros, A.~A., Shechtman, E., and Wang, O.
\newblock The unreasonable effectiveness of deep features as a perceptual
  metric.
\newblock In \emph{Proc. CVPR}, pp.\  586--595, 2018.

\end{thebibliography}
\bibliographystyle{icml2020}

\clearpage
\appendix

\section{Architecture}
\label{app:architecture}
For the \textbf{feature extractor} $\featureextractor$, we use the \emph{ResNet50 v2} architecture \cite{he2016identity, kolesnikov2019revisiting} with the standard channel widening factor of 4 (i.e. $16\times4$ channels in the first convolutional layer) and a representation size of 2048 at the \emph{pre-logits} layer unless otherwise noted.

\begin{figure*}[t]
    \centering
    \includegraphics[width=\textwidth]{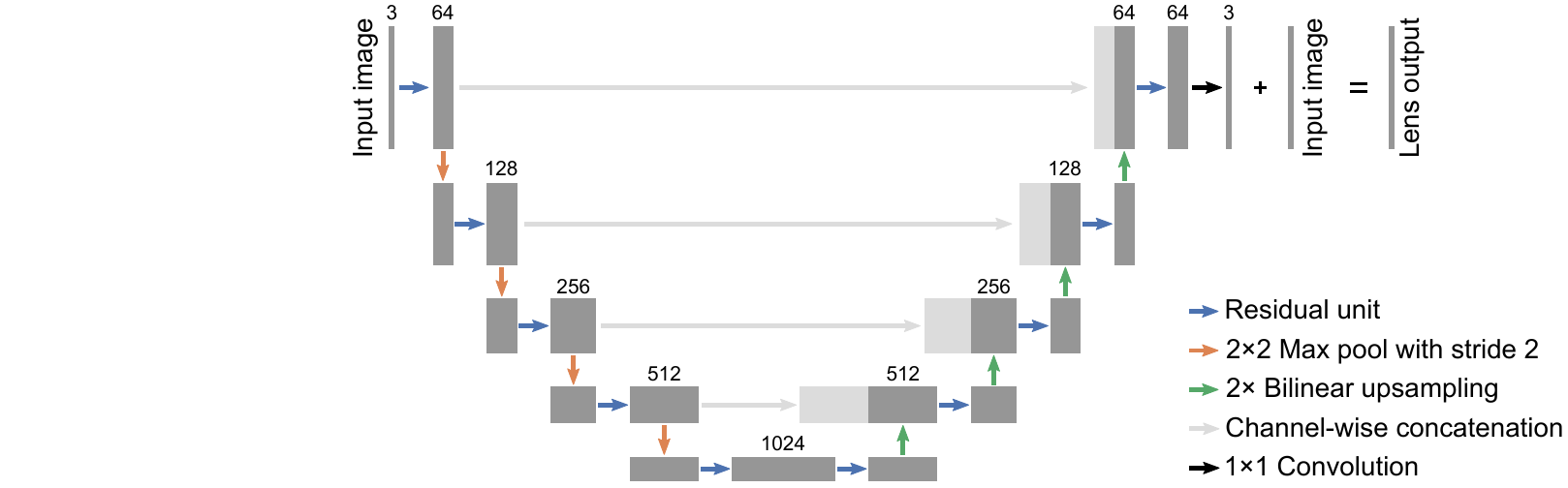}
    \caption{Lens architecture. The number of channels is indicated above each block. Based on \cite{ronneberger2015unet}.}
    \label{fig:unet_schematic}
\end{figure*}

For the \textbf{lens} $\lens$, we use a variant of the \emph{U-Net} architecture (Figure~\ref{fig:unet_schematic}; \citealt{ronneberger2015unet}). The lens consists of a convolutional encoder and decoder. The encoder and decoder are each a stack of $n$ residual units (same unit architecture as used for the feature extractor), with $k$ channels for the first unit of the encoder. We use $n=4$ and $k=64$ for all experiments. Two additional residual units form the bottleneck between encoder and decoder (see Figure~\ref{fig:unet_schematic}). After each unit in the encoder, the number of channels is doubled and the resolution is halved by max-pooling with a $2\times2$ kernel and stride 2. Conversely, after each decoder unit, the number of channels is halved and the resolution is doubled using bilinear interpolation. At each resolution level, skip connections are created between the encoder and decoder by concatenating the encoder representation channel-wise with the decoder representation before applying the next decoder unit. The output of the decoder is of the same resolution as the input image, and reduced to three channels by a $1\times1$ convolutional layer. This map is combined by element-wise addition with the input image to produce the lens output.

We choose the \emph{U-Net} architecture because it efficiently combines a large receptive field with a high output resolution. For example, for input images of size $224\times224$, the maps at the bottleneck of the \emph{U-Net} are of size $14\times14$, such that a $3\times3$ convolution at that size corresponds to $48\times48$ pixels at the input resolution and is able to capture large-scale image context. Furthermore, the skip connections of the \emph{U-Net} make it trivial for the lens to reconstruct the input image by setting all internal weights to zero. This is important to ensure that the changes made by the lens to the image are not simply due to a lack of capacity.

We find that a lens with $n=4$ and $k=64$ yields good results in general, although initial experiments suggested that tuning the lens capacity individually for each pretext task and dataset may provide further gains. 

\begin{figure}[t]
\begin{center}
    \centerline{\includegraphics[width=2.533in]{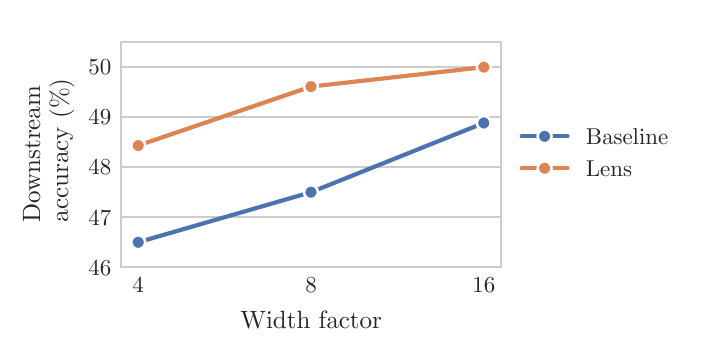}}
    \caption{Downstream accuracy for \rotation{} models trained on \imagenet{} with different feature extraction network widening factors. The performance gain remains large across model sizes.}
    \label{fig:model_width_sweep}
\end{center}
\vskip -0.2in
\end{figure}

We also tested how the performance of the lens varies with the capacity of the feature extraction network. For the \rotation{} task and \imagenet{}, we trained models with different widening factors (channel number multiplier). As expected, wider networks perform better (Figure~\ref{fig:model_width_sweep}). We find that the lens improves accuracy across all model widths. The accuracy gain of applying the lens to a feature extraction network with a width factor of 4 is equivalent to the gain obtained by widening the network by 2--$4\times$.

For the experiments using \cifar{} (Figure~\ref{fig:cifar_synthetic_shortcuts}), we used a smaller lens architecture consisting of a stack of five ResNet50 v2 residual units without down or up-sampling. 


\section{Downstream evaluation}
\label{app:evaluation}

\begin{table*}[t]\small
    \caption{Evaluation of representations from models trained on \imagenet{} with different self-supervised pretext tasks, using lensed-image representations only, without concatenating non-lensed representations. Otherwise like Table~\ref{table:main_results_table}: The scores are accuracies (in \%) of a logistic regression model trained on representations obtained from the frozen models. Mean $\pm$ s.e.m over three random initializations. Values in bold are better than the next-best method at a significance level of 0.05. Training images are preprocessed as suggested by the respective original works.}
    \label{table:results_table_without_concatenation}
    \vspace{0.2cm}
    \centering
    \begin{tabular}{llllll}
    \toprule
    Dataset & Method & \multicolumn{4}{c}{Pretext task} \\
    \cmidrule[0.5pt]{3-6}
              & &                            Rotation &                            Exemplar &                     Rel. patch loc. &                              Jigsaw \\
    \midrule
    ImageNet & Baseline &                     $45.9 \pm 0.04$ &                     $42.2 \pm 0.27$ &                     $37.5 \pm 0.17$ &                     $34.6 \pm 0.10$ \\
              & Lens &  $46.9 \pm 0.09$ $(\mathbf{+1.06})$ &  $44.5 \pm 0.12$ $(\mathbf{+2.26})$ &  $39.1 \pm 0.13$ $(\mathbf{+1.63})$ &  $38.2 \pm 0.09$ $(\mathbf{+3.63})$ \\
    
    \midrule
    Places205 & Baseline &                     $41.3 \pm 0.13$ &                     $41.8 \pm 0.15$ &                     $40.2 \pm 0.09$ &                     $38.8 \pm 0.21$ \\
              & Lens &  $41.8 \pm 0.14$ $(\mathbf{+0.53})$ &         $42.4 \pm 0.20$ $({+0.60})$ &  $40.9 \pm 0.05$ $(\mathbf{+0.70})$ &  $40.5 \pm 0.11$ $(\mathbf{+1.74})$ \\
    \bottomrule
    \end{tabular}
    \vspace{-0.2cm}
\end{table*}
\vspace{4mm}

For downstream evaluation of learned representations, we follow the linear evaluation protocol with SGD from \citet{kolesnikov2019revisiting}. A logistic regression model for \imagenet{} or \places{} classification was trained using SGD on the representations obtained from the pre-trained self-supervised models.

For training the logistic regression, we preprocessed input images in the same way for all models: Images were resized to $256\times256$, randomly cropped to $224\times224$, and the color values were scaled to $[-1, 1]$. For evaluation, the random crop was replaced by a central crop.

Representations were then obtained by passing the images through the pre-trained models and extracting the \emph{pre-logits} activations. For patch-based models, we obtained representations of the full image by averaging the representations of nine patches created from the full image. To create the patches, the the central $192\times192$ section of the $224\times224$ input image was divided into a $3\times3$ grid of patches. Each patch was passed through the feature extraction network and the representations were averaged.

The logistic regression model was trained with a batch size of 2048 and an initial learning rate of 0.8. We trained for 90 epochs and reduced the learning rate by a factor of 10 after epoch 50 and epoch 70. For both \imagenet{} and \places{}, training was performed on the full training set and the performance is reported for the public validation set.

\section{Adversarial training with FGSM}
\label{app:fgsm}
For the comparison to adversarial training (Table~\ref{table:main_results_table}), we used the fast gradient-sign method (FGSM) as described by \citet{kurakin2016adversarial}. Analogously to our sweeps over $\reclossscale$ for the lens models, we swept over the perturbation scale $\epsilon \in \{0.01, 0.02, 0.04, 0.08, 0.16\}$ and report the accuracy for the best $\epsilon$ in Table~\ref{table:main_results_table}. As suggested by \citet{kurakin2016adversarial}, we randomized the perturbation scale for each image by using the absolute value of a sample from a truncated normal distribution with mean 0 and standard deviation $\epsilon$. Since this randomization already includes nearly unprocessed images, we do not include further unprocessed images during training.

\section{Case study: SimCLR}
\label{sec:simclr}

Concurrently with our work, a powerful new self-supervised approach based on contrastive learning, called \emph{SimCLR}, was published \cite{chen2020simple}. Here, we describe our experience applying automatic shortcut removal to \emph{SimCLR} as an informal ``case study''. Our goal is to provide a practical example for how our method can be applied to understand and improve new self-supervised tasks. Even though we find that the lens does not improve the linear evaluation performance of \emph{SimCLR}, the lens provided insights that allowed us to improve SimCLR performance on other tasks.

\subsection{Linear evaluation on \imagenet{}}
\begin{figure}[ht]
\begin{center}
    \centerline{\includegraphics[width=\columnwidth]{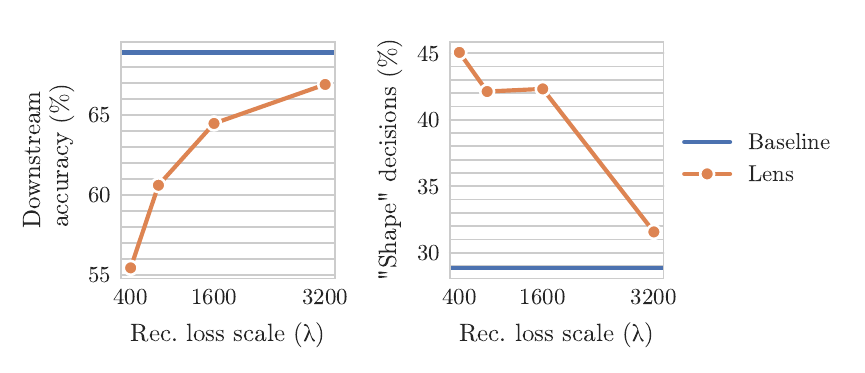}}
    \vspace{-5mm}
    \caption{\textbf{Left:} Linear evaluation performance of \emph{SimCLR} on \imagenet{}. \textbf{Right:} Fraction of ``shape'' decisions on the conflict stimuli from \citet{geirhos2019texture}}
    \label{fig:simclr_linear_accuracy}
\end{center}
\end{figure}

As a first step, we applied automatic shortcut removal as described in the main paper to \emph{SimCLR}\footnote{Code available at https://github.com/google-research/simclr; we used SimCLRv1.} and evaluated the learned representations with the linear protocol. As we suggest in the main paper, we ran a sweep across the reconstruction loss scale $\lambda$ and left the other hyperparameters at their default values. Figure~\ref{fig:simclr_linear_accuracy} shows that applying the lens to \emph{SimCLR} does not improve representation quality under the linear evaluation protocol. The performance increases monotonically with $\lambda$ and always remains below the baseline performance of 68.90 \% (\emph{ResNet50x1}), suggesting that any amount of lens-induced perturbation is harmful for this task under the linear evaluation protocol. To understand this result, we turned to inspecting the lens outputs.

\subsection{Lens outputs}
\begin{figure}[ht]
\begin{center}
    \centerline{\includegraphics[width=\columnwidth]{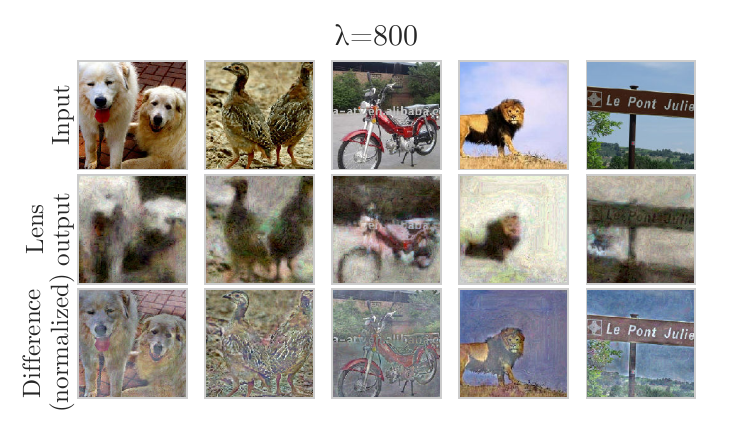}}
    \centerline{\includegraphics[width=\columnwidth]{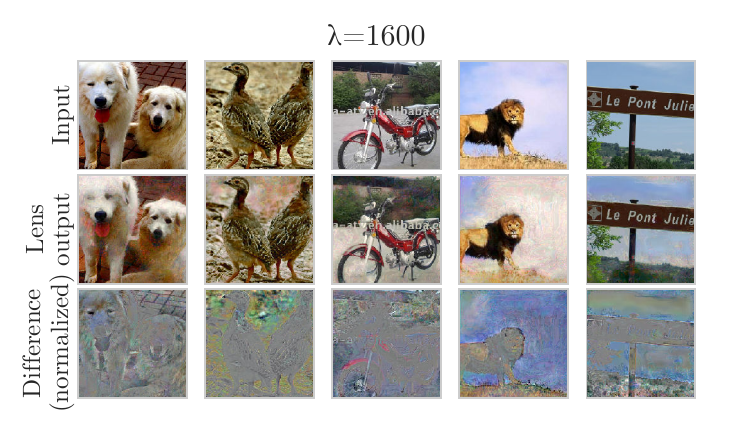}}
    \vspace{-5mm}
    \caption{Example lens outputs for \emph{SimCLR}.}
    \label{fig:simclr_example_images}
\end{center}
\end{figure}

The lens outputs (Figure~\ref{fig:simclr_example_images}) indicate that the lens primarily reduces color saturation and causes blurring of high-frequency image components. This suggests that the lens attacks features in a way that is similar to the augmentations that are part of the standard \emph{SimCLR} code, specifically \emph{Gaussian blur} and \emph{Color jitter}. These augmentations are an integral part of \emph{SimCLR}. We hypothesize that the augmentations were already so highly optimized that any additional image perturbation leads to a decrease in performance. Consistently, in separate experiments, we found that if we ablate the \emph{Gaussian blur} and \emph{Color jitter} augmentations, applying the lens improves over the ablated baseline (but not beyond the un-ablated baseline performance). While the lens does not provide further improvements on top of the hand-designed augmentations, it is encouraging that the lens identifies the same perturbations that were chosen by the expert authors of \emph{SimCLR}. 

\subsection{Semanticity}
The lens output suggests that high-frequency patterns, as well as colors, are important shortcut features for \emph{SimCLR}. We therefore hypothesized that the representations learned by \emph{SimCLR} primarily encode texture details, rather than high-level shape information. Indeed, evaluating \emph{SimCLR} on the dataset from \citet{geirhos2019texture} as in Section~\ref{subsec:results}, showed that \emph{SimCLR} makes shape-based decisions in only 28.86\% of cases (Figure~\ref{fig:simclr_linear_accuracy}). Applying the lens to \emph{SimCLR} increases the proportion of shape-based decisions to over 40\%, which indicates that the lens strongly shifts the network towards more semantic representations. 

\subsection{Improvements on other tasks}
While it has been shown that natural image classification tasks such as \imagenet{} classification can be solved accurately based on texture information \cite{geirhos2019texture}, other tasks might benefit from the additional semantic information that is learned when the lens is used. To investigate this question, we turned to the \emph{Visual Task Adaptation Benchmark} (\emph{VTAB}, \citealt{zhai2019vtab}), which is a collection of 19 tasks that span \emph{natural}, \emph{specialized} and \emph{structured} domains. Indeed, we find that automatic shortcut removal improves the mean score of \emph{SimCLR} on \emph{VTAB} by 2.72\% (Table~\ref{table:simclr_vtab}). This improvement comes primarily from the Specialized and Structured datasets, while the score on Natural datasets is slightly reduced. These results suggest that \emph{SimCLR} representations are highly adapted to \imagenet{}, and their performance on a wider variety of tasks may suffer from shortcuts that can be mitigated with our method.

\begin{table}[t]\small
    \caption{Fine-tuning performance of \emph{SimCLR R50x1} on the \emph{Visual Task Adaptation Benchmark} \cite{zhai2019vtab}. Abbreviations: Spec., Specialized; Struct., Structured.}
    \label{table:simclr_vtab}
    \vspace{0.2cm}
    \centering
    \small
    \begin{tabular}{lllll}
    \toprule
      & mean & Natural & Spec. & Struct. \\
    \midrule
    Baseline & 48.79 & 51.08 & 74.74 & 33.82 \\
    Lens     & 51.51 \textbf{(+2.72})  & 49.96 & 76.89 & 40.17 \\
    \bottomrule
    \end{tabular}
\end{table}

\subsection{Summary}
The \emph{SimCLR} case study shows how our method can be used to understand and improve a new pretext task. While our method does not always result in a quick win on all benchmarks, it provides a deeper understanding of the task-specific shortcut features, which may guide the practitioner towards opportunities for improvement.

\begin{figure*}[t]
\begin{center}
    \includegraphics[width=\textwidth,trim={0.1in 0.1in 0.1in 0.1in},clip]{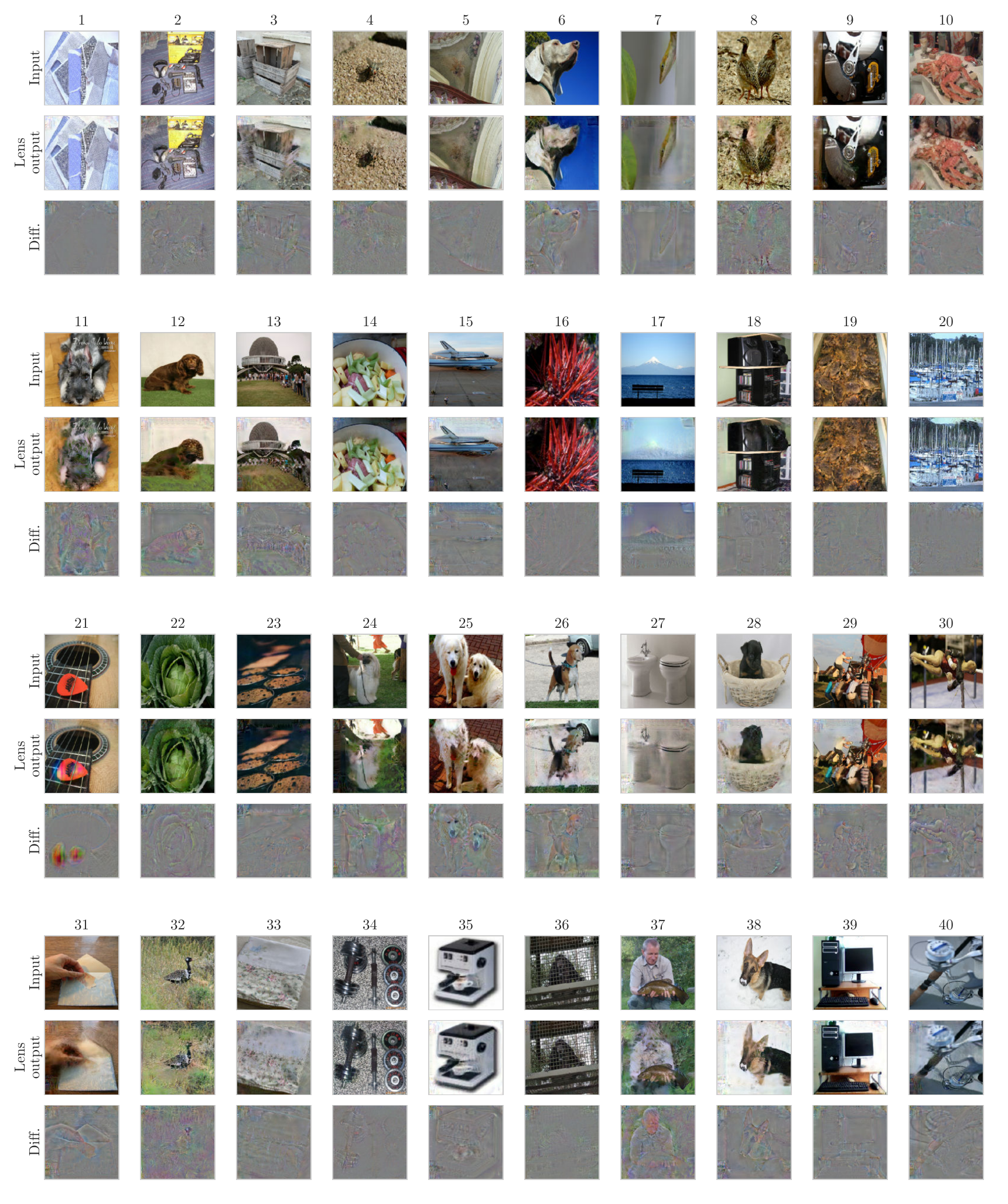}
    \caption{Further example lens outputs for models trained on \imagenet{} with the \rotation{} task. Images were randomly sampled from the \imagenet{} validation set.}
\end{center}
\end{figure*}

\begin{figure*}[t]
\begin{center}
    \includegraphics[width=\textwidth,trim={0.1in 0.1in 0.1in 0.1in},clip]{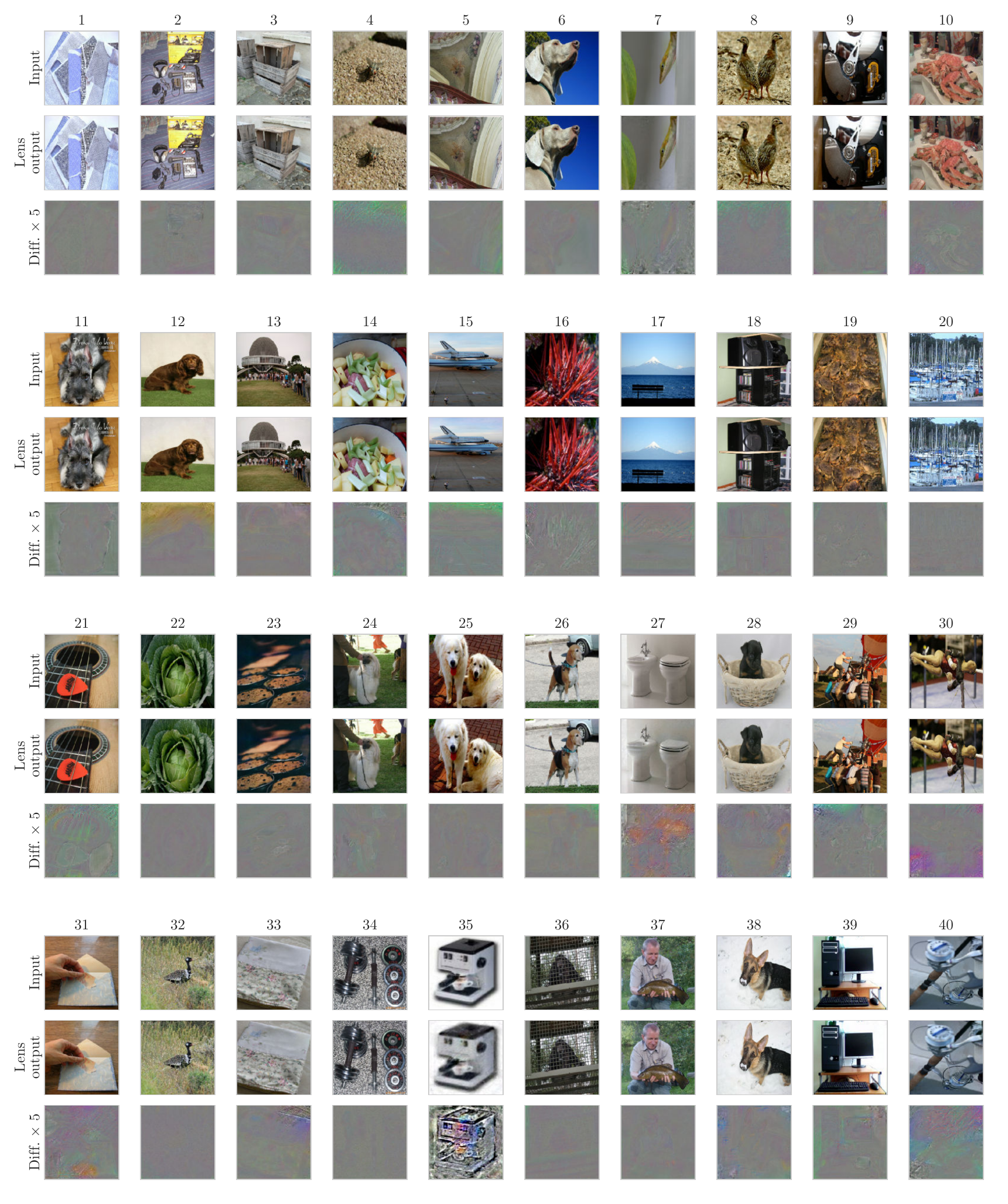}
    \caption{Further example lens outputs for models trained on \imagenet{} with the \exemplar{} task. Images were randomly sampled from the \imagenet{} validation set.}
\end{center}
\end{figure*}

\begin{figure*}[t]
\begin{center}
    \includegraphics[width=\textwidth,trim={0.1in 0.1in 0.1in 0.1in},clip]{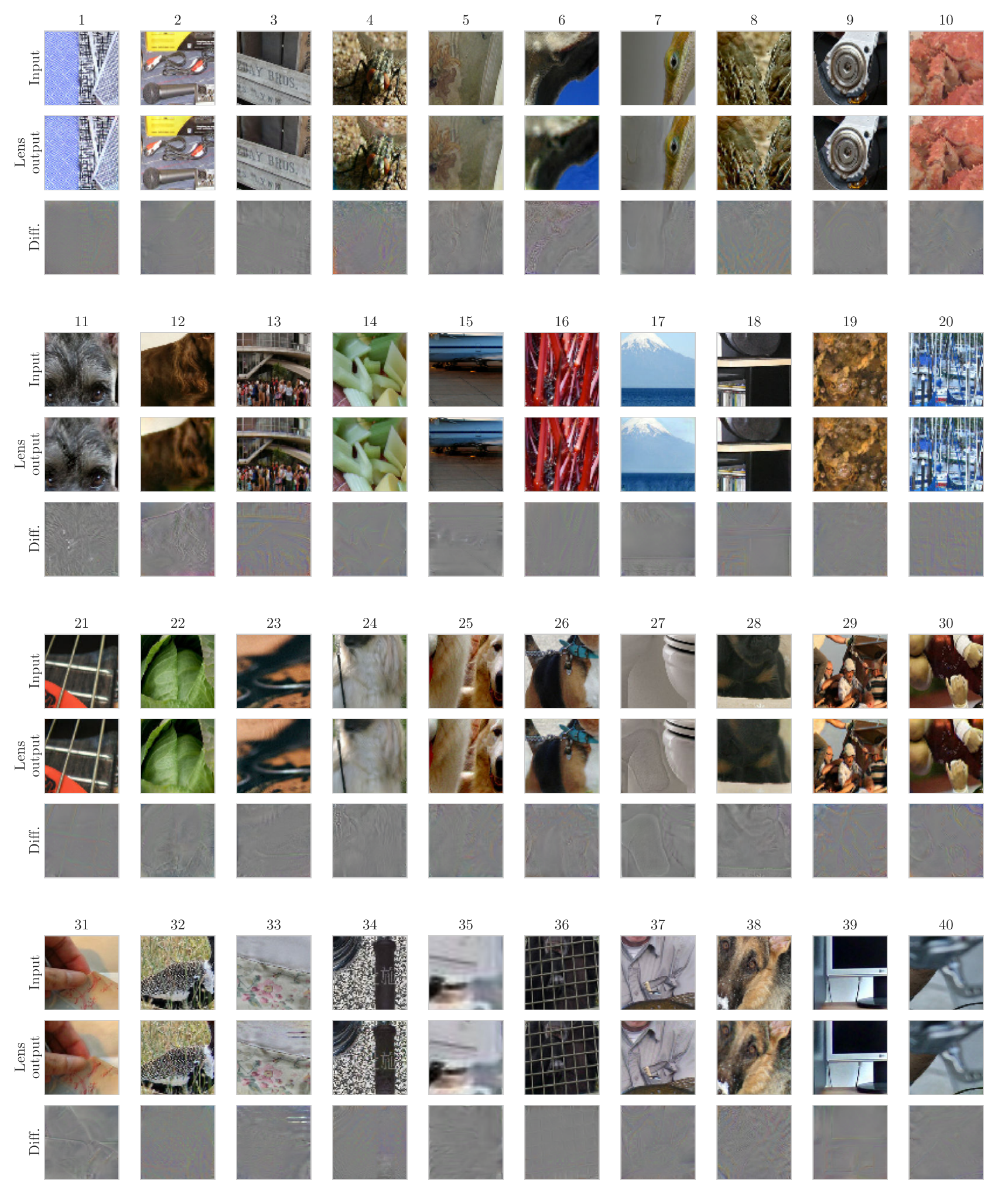}
    \caption{Further example lens outputs for models trained on \imagenet{} with the \relpatchloc{} task. Images were randomly sampled from the \imagenet{} validation set.}
\end{center}
\end{figure*}

\begin{figure*}[t]
\begin{center}
    \includegraphics[width=\textwidth,trim={0.1in 0.1in 0.1in 0.1in},clip]{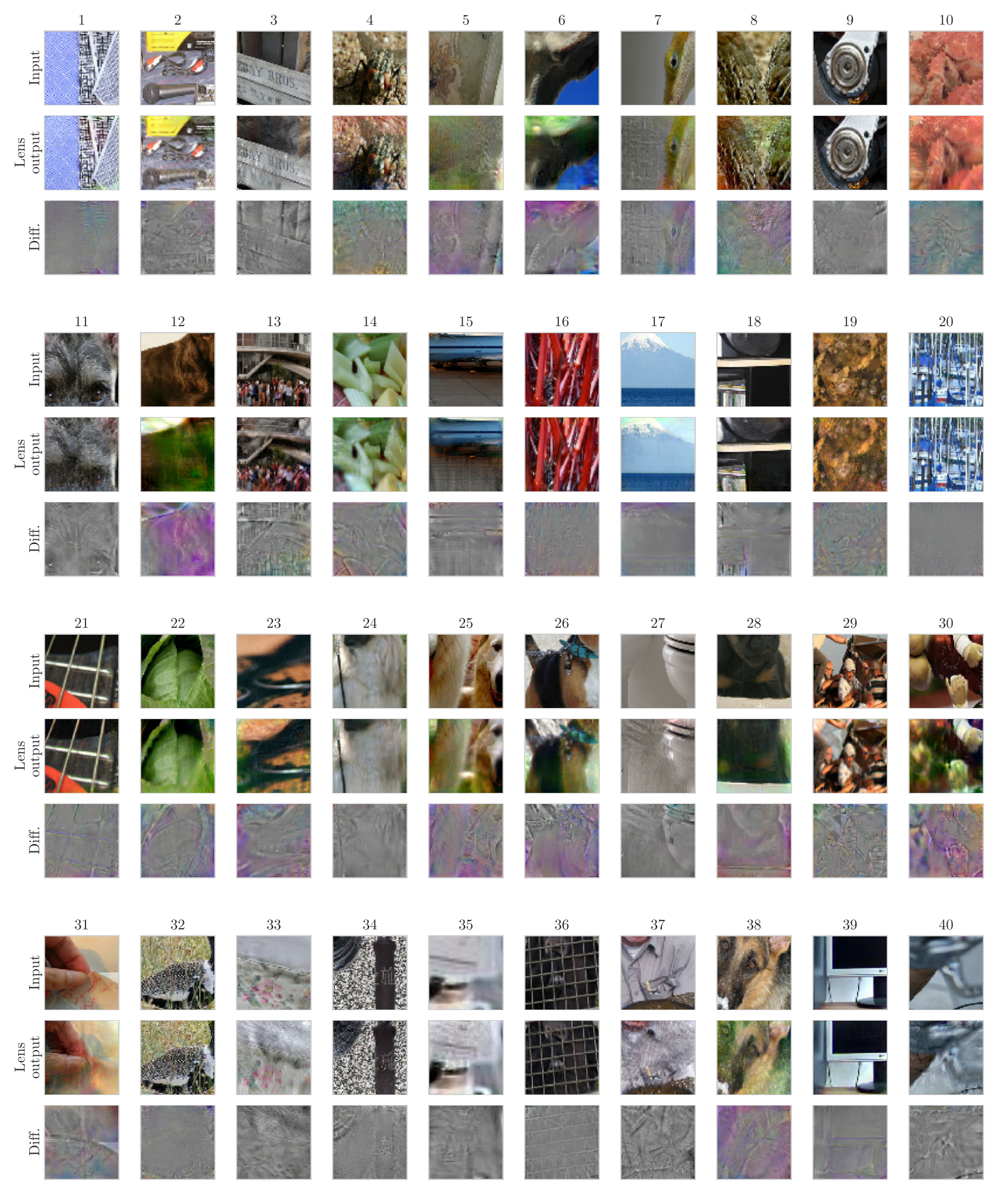}
    \caption{Further example lens outputs for models trained on \imagenet{} with the \jigsaw{} task. Images were randomly sampled from the \imagenet{} validation set.}
\end{center}
\end{figure*}

\begin{figure*}[t]
\begin{center}
    \includegraphics[width=\textwidth,trim={0.1in 0.1in 0.1in 0.1in},clip]{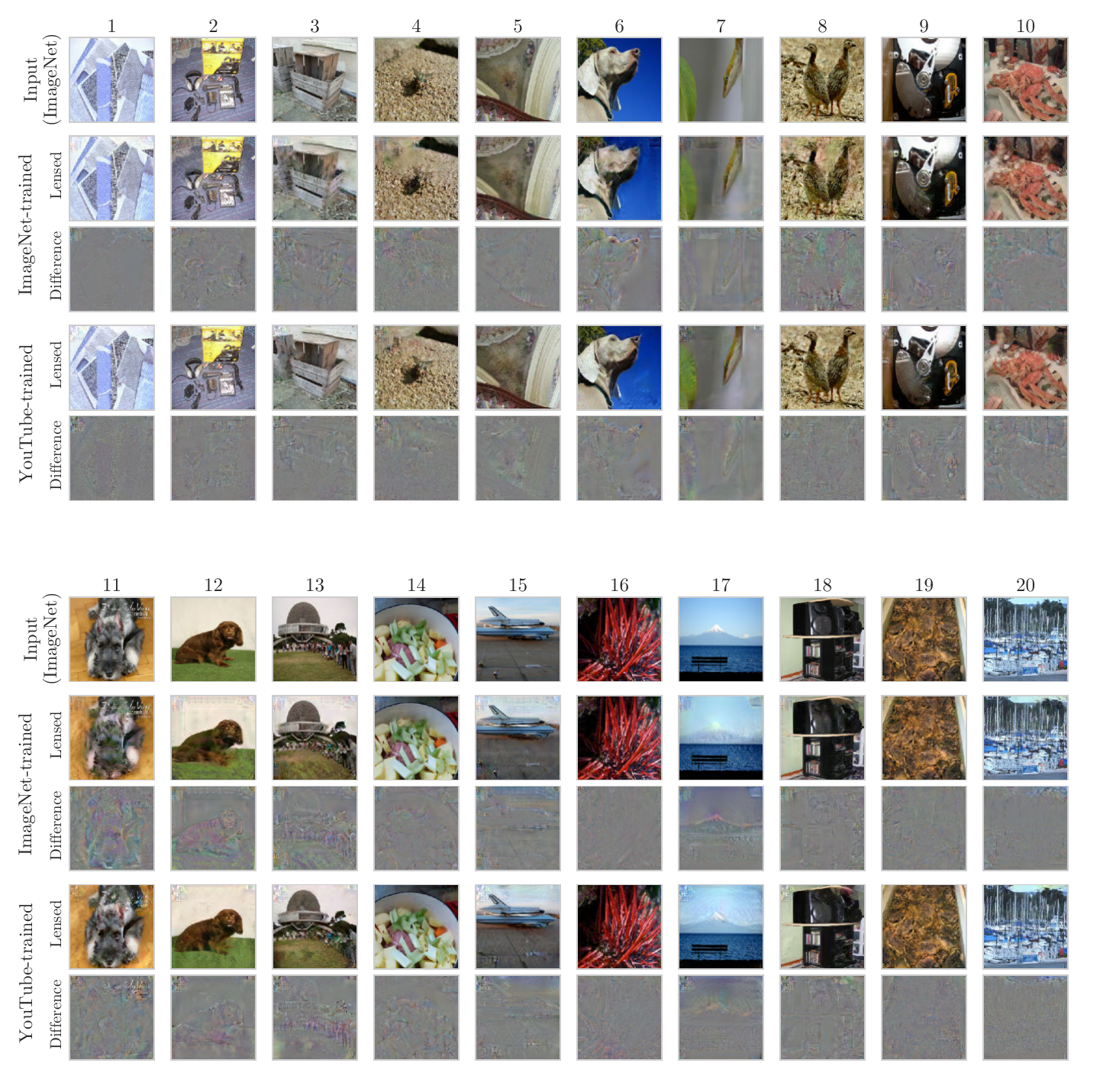}
    \caption{Further example lens outputs for models trained on \youtube{} with the \rotation{} task. Outputs from \imagenet{}-trained models are provided for comparison. Images were randomly sampled from the \imagenet{} validation set.}
\end{center}
\end{figure*}

\begin{figure*}[t]
\begin{center}
    \includegraphics[width=\textwidth,trim={0.1in 0.1in 0.1in 0.1in},clip]{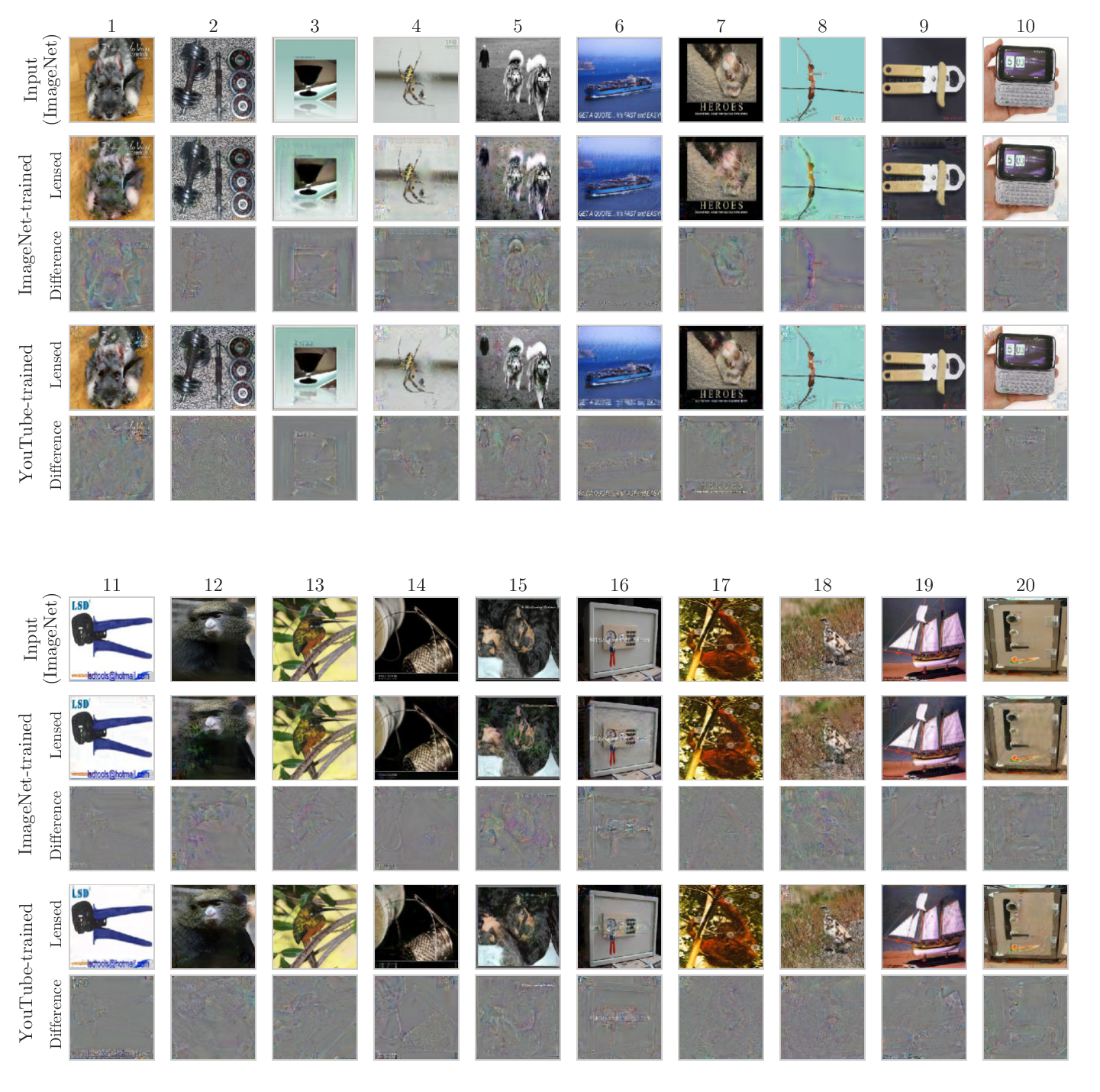}
    \caption{Further example lens outputs for images containing text, comparing models trained on \youtube{} and \imagenet{} with the \rotation{} task. Images containing artificially overlaid text (logos, watermarks, etc.) were manually selected from a random sample of 1000 \imagenet{} validation images, before inspecting lens outputs. A random sample of these images is shown.}
\end{center}
\end{figure*}

\end{document}